# Conditional Noise-Contrastive Estimation of Unnormalised Models

Ciwan Ceylan [* 1]  Michael U. Gutmann [* 2]

## Abstract

Many parametric statistical models are not properly normalised and only specified up to an intractable partition function, which renders parameter estimation difficult. Examples of unnormalised models are Gibbs distributions, Markov random fields, and neural network models in unsupervised deep learning. In previous work, the estimation principle called noise-contrastive estimation (NCE) was introduced where unnormalised models are estimated by learning to distinguish between data and auxiliary noise. An open question is how to best choose the auxiliary noise distribution. We here propose a new method that addresses this issue. The proposed method shares with NCE the idea of formulating density estimation as a supervised learning problem but in contrast to NCE, the proposed method leverages the observed data when generating noise samples. The noise can thus be generated in a semi-automated manner. We first present the underlying theory of the new method, show that score matching emerges as a limiting case, validate the method on continuous and discrete valued synthetic data, and show that we can expect an improved performance compared to NCE when the data lie in a lower-dimensional manifold. Then we demonstrate its applicability in unsupervised deep learning by estimating a four-layer neural image model.

## 1. Introduction

We consider the problem of estimating the parameters $\boldsymbol{\theta} \in \mathbb{R}^M$ of an unnormalised statistical model $\phi(\mathbf{u}; \boldsymbol{\theta}) : \mathbb{X} \mapsto \mathbb{R}^+$ from observed data $\mathbf{X} = \{\mathbf{x}_1, \ldots, \mathbf{x}_N\}$, where the

$\mathbf{x}_i \in \mathbb{X}$ are independently sampled from the unknown data distribution $p_d$. Unnormalised models output non-negative numbers but do not integrate or sum to one, i.e. they are statistical models that are defined up to the partition function $Z(\boldsymbol{\theta}) = \int \phi(\mathbf{u}; \boldsymbol{\theta}) \, d\mathbf{u}$. Unnormalised models are widely used, e.g. to model images (Köster & Hyvärinen, 2010; Gutmann & Hyvärinen, 2013), natural language (Mnih & Teh, 2012; Zoph et al., 2016), or memory (Hopfield, 1982).

If the partition function $Z(\boldsymbol{\theta})$ can be evaluated analytically in closed form, the unnormalised model $\phi(\mathbf{u}; \boldsymbol{\theta})$ can be easily converted to a (normalised) statistical model $p(\mathbf{u}; \boldsymbol{\theta}) = \phi(\mathbf{u}; \boldsymbol{\theta})/Z(\boldsymbol{\theta})$ that can be estimated by maximising the likelihood. However, for most unnormalised models the integral defining the partition function is analytically intractable and computationally expensive to approximate.

Several methods have been proposed in the literature to estimate unnormalised models including Monte Carlo maximum likelihood (Geyer, 1994), contrastive divergence (Hinton, 2002), score matching (Hyvärinen, 2005), and noise-contrastive estimation (Gutmann & Hyvärinen, 2010; 2012) and its generalisations (Pihlaja et al., 2010; Gutmann & Hirayama, 2011). The basic idea of noise-contrastive estimation (NCE) is to formulate the density estimation problem as a classification problem where the model is trained to distinguish between the observed data and some reference (noise) data. NCE is used in several application domains (Mnih & Teh, 2012; Chen et al., 2015; Tschiatschek et al., 2016) and similar "learning by comparison" ideas are employed for learning with generative latent variable models (Gutmann et al., 2014; Goodfellow et al., 2014).

In NCE, the choice of the auxiliary noise distribution is left to the user. While simple distributions, e.g. uniform or Gaussian distributions, have successfully been used (Gutmann & Hyvärinen, 2012; Mnih & Teh, 2012), the estimation performance of NCE depends on the distribution chosen and more tailored distributions were found to typically yield better results, see e.g. (Ji et al., 2016). Intuitively, the noise samples in NCE ought to resemble the observed data in order for the classification problem not to be too easy. To alleviate the burden on the user to generate such noise, we here propose *conditional noise-contrastive estimation* that semi-automatically generates the noise based on the observed data.

---

*Equal contribution  [1]UMIC, RWTH Aachen University, Aachen, Germany (affiliated with KTH Royal Institute of Technology and University of Edinburgh during project timespan) [2]School of Informatics, University of Edinburgh, Edinburgh, United Kingdom. Correspondence to: Ciwan Ceylan <ceylan@vision.rwth-aachen.de>, Michael Gutmann <michael.gutmann@ed.ac.uk>.

The rest of the paper is structured as follows. In Section 2, we present the theory of conditional noise-contrastive estimation (CNCE), establish basic properties, and prove that a limiting case yields score matching. In Section 3, we validate the theory on synthetic data and compare the estimation performance of CNCE with NCE. In Section 4, we apply CNCE to real data and show that it can handle complex models by estimating a four-layer neural network model of natural images, and Section 5 concludes the paper.

## 2. Conditional noise-contrastive estimation

Conditional noise-contrastive estimation (CNCE) turns an unsupervised estimation problem into a supervised learning problem by training the model to distinguish between data and noise samples. This is the same high-level approach as NCE takes, but in contrast to NCE, the novel idea of CNCE is to generate the noise samples with the aid of the observed data samples. Therefore, unlike NCE, CNCE does not assume the noise samples to be generated independently of the data samples, but rather to be drawn from a conditional noise distribution $p_c$. The generated noise samples are paired with the data samples, with $\kappa$ noise samples $\mathbf{y}_{ij} \in \mathbb{Y}, j = 1, \ldots, \kappa$ per observed data point $\mathbf{x}_i$. Thus, a total of $N \cdot \kappa$ noise samples $\mathbf{y}_{ij} \sim p_c(\mathbf{y}_{ij}|\mathbf{x}_i)$ are generated from $p_c$. We denote the collection of all noise samples by $\mathbf{Y}$. In what follows, we assume that $\mathbb{X} = \mathbb{Y}$, but this assumption can be relaxed to $\mathbb{X} \subseteq \mathbb{Y}$ (see Supplementary Materials A). In any case, we denote the union of $\mathbb{X}$ and $\mathbb{Y}$ by $\mathbb{U}$.

We derive the loss function for CNCE in analogy to the derivation of the loss function for NCE. We divide all pairs of data and noise samples into two classes, $C_{\boldsymbol{\alpha}}$ and $C_{\boldsymbol{\beta}}$, of equal size. Class $C_{\boldsymbol{\alpha}}$ is formed by tuples $(\mathbf{u}_1, \mathbf{u}_2)$ with $\mathbf{u}_1 \in \mathbf{X}$ and $\mathbf{u}_2 \in \mathbf{Y}$, while $C_{\boldsymbol{\beta}}$ is formed by tuples $(\mathbf{u}_1, \mathbf{u}_2)$ with $\mathbf{u}_1 \in \mathbf{Y}$ and $\mathbf{u}_2 \in \mathbf{X}$. Consequently, the probability distributions for the classes $C_{\boldsymbol{\alpha}}$ and $C_{\boldsymbol{\beta}}$ are given by

$$p_{\boldsymbol{\alpha}}(\mathbf{u}_1, \mathbf{u}_2) = p_d(\mathbf{u}_1)p_c(\mathbf{u}_2|\mathbf{u}_1), \tag{1}$$

$$p_{\boldsymbol{\beta}}(\mathbf{u}_1, \mathbf{u}_2) = p_d(\mathbf{u}_2)p_c(\mathbf{u}_1|\mathbf{u}_2), \tag{2}$$

where $p_d$ denotes the distribution of the $\mathbf{x}_i$. The class conditional distributions can be obtained by Bayes' rule,

$$p_{C_{\boldsymbol{\alpha}}|\mathbf{u}}(\mathbf{u}_1, \mathbf{u}_2) = \frac{p_{\boldsymbol{\alpha}}(\mathbf{u}_1, \mathbf{u}_2)}{p_{\boldsymbol{\alpha}}(\mathbf{u}_1, \mathbf{u}_2) + p_{\boldsymbol{\beta}}(\mathbf{u}_1, \mathbf{u}_2)} \tag{3}$$

$$= \frac{1}{1 + \frac{p_d(\mathbf{u}_2)p_c(\mathbf{u}_1|\mathbf{u}_2)}{p_d(\mathbf{u}_1)p_c(\mathbf{u}_2|\mathbf{u}_1)}}, \tag{4}$$

$$p_{C_{\boldsymbol{\beta}}|\mathbf{u}}(\mathbf{u}_1, \mathbf{u}_2) = \frac{1}{1 + \frac{p_d(\mathbf{u}_1)p_c(\mathbf{u}_2|\mathbf{u}_1)}{p_d(\mathbf{u}_2)p_c(\mathbf{u}_1|\mathbf{u}_2)}}. \tag{5}$$

The prior class probabilities cancel because there are equally many samples in each class.

By replacing $p_d(\cdot)$ with the model $\phi(\cdot; \boldsymbol{\theta})/Z(\boldsymbol{\theta})$, the partition functions cancel and the following parametrised ver-

sions of the class conditional distributions are obtained

$$p_{C_{\boldsymbol{\alpha}}|\mathbf{u}}(\mathbf{u}_1, \mathbf{u}_2; \boldsymbol{\theta}) = \frac{1}{1 + \frac{\phi(\mathbf{u}_2; \boldsymbol{\theta})p_c(\mathbf{u}_1|\mathbf{u}_2)}{\phi(\mathbf{u}_1; \boldsymbol{\theta})p_c(\mathbf{u}_2|\mathbf{u}_1)}}, \tag{6}$$

$$p_{C_{\boldsymbol{\beta}}|\mathbf{u}}(\mathbf{u}_1, \mathbf{u}_2; \boldsymbol{\theta}) = \frac{1}{1 + \frac{\phi(\mathbf{u}_1; \boldsymbol{\theta})p_c(\mathbf{u}_2|\mathbf{u}_1)}{\phi(\mathbf{u}_2; \boldsymbol{\theta})p_c(\mathbf{u}_1|\mathbf{u}_2)}}. \tag{7}$$

The CNCE loss function is now formed as the negative log likelihood over the conditional class probabilities, in the same manner as in NCE (Gutmann & Hyvärinen, 2012),

$$\mathcal{J}_N(\boldsymbol{\theta}) = \frac{2}{\kappa N} \sum_{j=1}^{\kappa} \sum_{i=1}^{N} \log\left[1 + \exp(-G(\mathbf{x}_i, \mathbf{y}_{ij}; \boldsymbol{\theta}))\right], \tag{8}$$

$$G(\mathbf{u}_1, \mathbf{u}_2; \boldsymbol{\theta}) = \log \frac{\phi(\mathbf{u}_1; \boldsymbol{\theta})p_c(\mathbf{u}_2|\mathbf{u}_1)}{\phi(\mathbf{u}_2; \boldsymbol{\theta})p_c(\mathbf{u}_1|\mathbf{u}_2)}. \tag{9}$$

The CNCE loss function $\mathcal{J}_N$ is the sample version of $\mathcal{J}(\boldsymbol{\theta}) = 2\mathbb{E}_{\mathbf{xy}} \log (1 + \exp(-G(\mathbf{x}, \mathbf{y}; \boldsymbol{\theta})))$, which is obtained by taking both $N$ and $\kappa$ to the $\infty$ limit. To further develop the theory, it is helpful to write $\mathcal{J}(\boldsymbol{\theta})$ as a functional of $G$, which gives

$$\tilde{\mathcal{J}}[G] = 2\mathbb{E}_{\mathbf{xy}} \log (1 + \exp(-G(\mathbf{x}, \mathbf{y}))). \tag{10}$$

We then obtain the following theorem:

**Theorem (Nonparametric estimation).** *Let* $G : \mathbb{U} \times \mathbb{U} \to \mathbb{R}$ *be a function of the form*

$$G(\mathbf{u}_1, \mathbf{u}_2) = f(\mathbf{u}_1) - f(\mathbf{u}_2) + \log \frac{p_c(\mathbf{u}_2|\mathbf{u}_1)}{p_c(\mathbf{u}_1|\mathbf{u}_2)}, \tag{11}$$

*where* $f$ *is a function from* $\mathbb{U}$ *to* $\mathbb{R}$. *Under the assumption* $\mathbb{X} = \mathbb{Y}$, $\tilde{\mathcal{J}}$ *attains a unique minimum at*

$$G^*(\mathbf{u}_1, \mathbf{u}_2) = \log \frac{p_d(\mathbf{u}_1)p_c(\mathbf{u}_2|\mathbf{u}_1)}{p_d(\mathbf{u}_2)p_c(\mathbf{u}_1|\mathbf{u}_2)} \tag{12}$$

*for* $(\mathbf{u}_1, \mathbf{u}_2) \in \mathbb{X} \times \mathbb{X}$ *with* $p_d(\mathbf{u}_1) > 0$ *and* $p_c(\mathbf{u}_1|\mathbf{u}_2) > 0$.

The proof of a more general version is given in Supplementary Materials A. The theorem shows that in the limit of large $N$ and $\kappa$, the optimal function $f$ equals $p_d$ up to an additive constant. For parametrisations that are flexible enough so that $G(\mathbf{u}_1, \mathbf{u}_2; \boldsymbol{\theta}^*) = G^*(\mathbf{u}_1, \mathbf{u}_2)$ for some value $\boldsymbol{\theta}^*$, the theorem together with the definition of $G(\mathbf{u}_1, \mathbf{u}_2; \boldsymbol{\theta})$ in (9) implies that $\phi(\mathbf{u}; \boldsymbol{\theta}^*) \propto p_d(\mathbf{u})$. We have here the proportionality sign because the normalising constant is not estimated in CNCE.

While the theorem above concerns nonparametric estimation, and hence does not take into account how $G$ is parametrised, it forms the basis for a consistency proof of CNCE. A standard approach is to identify conditions under which $\mathcal{J}_N(\boldsymbol{\theta})$ converges uniformly in probability to $\mathcal{J}(\boldsymbol{\theta})$ and then to appeal to e.g. Theorem 5.7 of (van der Vaart, 1998). A similar approach where the Kullback-Leibler divergence takes the role of $\mathcal{J}$ can be used to prove consistency of maximum likelihood estimation. The conditions for uniform convergence are typically fairly technical and we here forego this endeavour and instead provide empirical evidence for consistency in Section 3.

The generic CNCE algorithm generally takes two steps: obtain the noise samples by sampling from the conditional noise distribution $p_c$, and then minimise the loss function $\mathcal{J}_N$ over the parameters $\boldsymbol{\theta}$. The user decides the trade-off between precision and computational expenditures via $\kappa$ and also needs to provide $p_c$.

There are two advantages to choosing $p_c$ over choosing the noise distribution in NCE. First, the observed data samples can be leveraged for sampling the noise, meaning that a resemblance to $p_d$ is easier to achieve than it would be for NCE. Indeed, all simulations in the paper were performed with the simple Gaussian specified below. Second, if $p_c$ is known to be symmetric, i.e. $p_c(\mathbf{u}_1|\mathbf{u}_2) = p_c(\mathbf{u}_2|\mathbf{u}_1)$, it does not need to be evaluated because the densities cancel out in Equation (9).

A simple symmetric choice of $p_c$ when $\mathbf{x}$ and $\mathbf{y} \in \mathbb{R}^D$ is

$$p_c(\mathbf{y}|\mathbf{x};\varepsilon) = \mathcal{N}(\mathbf{y};\mathbf{x},\varepsilon^2\mathbb{1}), \quad \mathbf{y}_{ij} = \mathbf{x}_i + \varepsilon\boldsymbol{\xi}_{ij}. \quad (13)$$

Here $\mathbb{1}$ is the identity matrix, $\boldsymbol{\xi}_{ij} \in \mathbb{R}^D$ is a multivariate standard normal random variable and $\varepsilon \in [0,\infty)$ a scalar parameter that corresponds to the standard deviation of each dimension, and which therefore controls the similarity between $\mathbf{Y}$ and $\mathbf{X}$. It is here assumed that the data have been standardised (Murphy, 2012, Chapter 4) so that the empirical variances of the data are one for each dimension. Otherwise, different values of $\varepsilon$ ought to be used for each dimension.

CNCE is also applicable to discrete random variables, e.g. by using a multinoulli distribution over $\mathbf{y}$ conditioned on $\mathbf{x}$, and non-negative data (see Supplementary Materials C).

In our simulations, we adjust $\varepsilon$ using simple heuristics so that the gradients of the loss function are not too small. This typically occurs when $\varepsilon$ is too large so that the noise and data are easily distinguishable, but also when $\varepsilon$ is too small. It can be verified that the loss function attains the value $2\log(2)$ for $\varepsilon = 0$ independent of the model and $\boldsymbol{\theta}$. In brief, the heuristic algorithm starts with a small $\varepsilon$ that is incremented until the value of the loss function is sufficiently far away from $2\log(2)$.

While small $\varepsilon$ cause the gradients to be small in absolute terms, the following theorem shows that the loss function remains meaningful and that CNCE then corresponds to score matching (Hyvärinen, 2005).

**Theorem (Connection to score matching).** *Assume that $\phi(\mathbf{u};\boldsymbol{\theta})$ is an unnormalised probability density and that*

*$f_{\boldsymbol{\theta}}(\mathbf{u}) = \log\phi(\mathbf{u};\boldsymbol{\theta})$ is twice differentiable. If $\mathbf{y} = \mathbf{x} + \varepsilon\boldsymbol{\xi}$ where $\boldsymbol{\xi}$ is a vector of uncorrelated random variables of mean zero and variance one that are independent from $\mathbf{x}$ and have a symmetric density, then*

$$\mathcal{J}(\boldsymbol{\theta}) = \frac{\varepsilon^2}{2}\mathbb{E}_{\mathbf{x}}\left[\sum_i \frac{\partial^2 f_{\boldsymbol{\theta}}(\mathbf{x})}{\partial x_i^2} + \frac{1}{2}||\nabla_{\mathbf{x}}f_{\boldsymbol{\theta}}(\mathbf{x})||_2^2\right] + 2\log(2) + O(\varepsilon^3).$$

The term in the brackets is the loss function that is minimised in score matching (Hyvärinen, 2005). The theorem is proved in Supplementary Materials B. Note that $p_c$ in (13) fulfills the conditions in the theorem.

The theorem can be understood as follows: Score matching consists in finding parameter values so that the slope of the model pdf matches the slope of the data pdf. For symmetric conditional noise distributions $p_c$, the nonlinearity $G$ in Equation (9) equals $G(\mathbf{u}_1,\mathbf{u}_2;\boldsymbol{\theta}) = \log\phi(\mathbf{u}_1;\boldsymbol{\theta}) - \log\phi(\mathbf{u}_2;\boldsymbol{\theta}) = f_{\boldsymbol{\theta}}(\mathbf{u}_1) - f_{\boldsymbol{\theta}}(\mathbf{u}_2)$. From (12), we know that at the optimum of $\mathcal{J}(\boldsymbol{\theta})$, $G(\mathbf{u}_1,\mathbf{u}_2;\boldsymbol{\theta})$ matches $\log p_d(\mathbf{u}_1) - \log p_d(\mathbf{u}_2)$. The values which the arguments $\mathbf{u}_1$ and $\mathbf{u}_2$ take during the minimisation are determined by the conditional noise distribution. For small $\varepsilon$, the arguments are always close to each other, so that $G(\mathbf{u}_1,\mathbf{u}_2;\boldsymbol{\theta})$ is approximately proportional to a directional derivative of $f_{\boldsymbol{\theta}}(\mathbf{u}) = \log\phi(\mathbf{u};\boldsymbol{\theta})$ along a random direction. This means that for small $\varepsilon$, $\mathcal{J}(\boldsymbol{\theta})$ is minimised when the slope of the model pdf matches the slope of the data pdf, as in score matching.

## 3. Empirical validation of the theory

We here validate consistency and compare CNCE with NCE on synthetic data. The models below were used in unnormalised form for CNCE and NCE. For the results with MLE, the models were first normalised. Additional results for non-negative and discrete data are provided in Supplementary Materials C.

### 3.1. Models

**The Gaussian model** is an unnormalised multivariate Gaussian model in five dimensions with zero mean and parametrised precision matrix $\boldsymbol{\Lambda}$. As the precision matrix is symmetric, the Gaussian model has 15 parameters,

$$\log\phi(\mathbf{u};\boldsymbol{\Lambda}) = -\frac{1}{2}\mathbf{u}^T\boldsymbol{\Lambda}\mathbf{u}, \qquad \mathbf{u} \in \mathbb{R}^5. \quad (15)$$

The estimation error was measured as the Euclidean distance between the true and estimated parameters.

**The ICA model** is commonly used in signal possessing for blind source separation (Hyvärinen & Oja, 2000). Assuming equally many sources as data dimensions, $D = 4$, and a

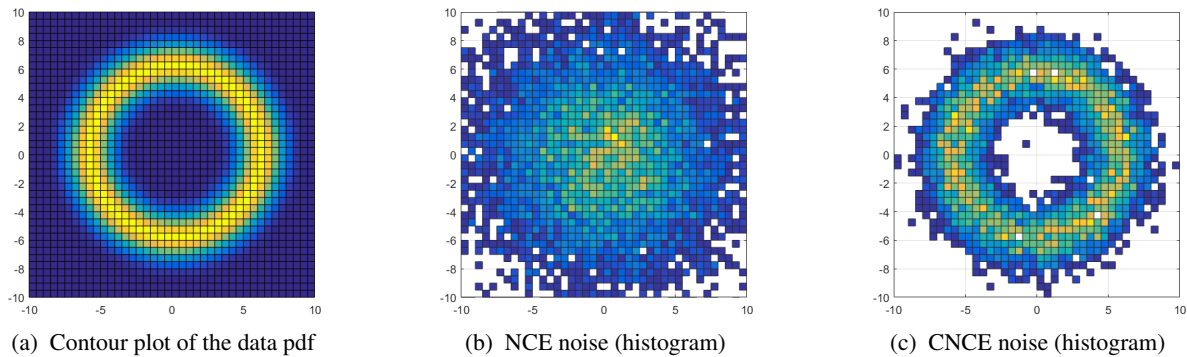

(a) Contour plot of the data pdf     (b) NCE noise (histogram)     (c) CNCE noise (histogram)

Figure 1: Visualisation of the ring model distribution and corresponding NCE and CNCE noise in two dimensions.

Laplacian distribution for the sources, the unnormalised ICA model is

$$\log \phi(\mathbf{u}; \mathbf{B}) = -\sqrt{2} \sum_{j=1}^{D} |\mathbf{b}_j \cdot \mathbf{u}|, \qquad \mathbf{u} \in \mathbb{R}^4. \quad (16)$$

The model is parametrised by the demixing matrix $\mathbf{B}$ and has $D^2 = 16$ free parameters. The (normalised) ICA model can be estimated using MLE (Hyvärinen & Oja, 2000, 4.4.1). The estimation error was calculated as the Euclidean distance between true and estimated parameter vector after accounting for the sign and order ambiguity of the ICA model (Hyvärinen & Oja, 2000, 2.2) in the same manner as in (Gutmann & Hyvärinen, 2012).

Both the Gaussian and the ICA model were previously used to validate the consistency of NCE, and a Gaussian noise distribution achieved good estimation performance (Gutmann & Hyvärinen, 2012). In order to investigate the potential benefit of the adaptive noise of CNCE, we used the following more challenging "ring model" where the data lie in lower dimensional manifold.

***The Ring model*** is given by

$$\log \phi(\mathbf{u}; \mu_r, \gamma_r) = -\frac{\gamma_r}{2} (\|\mathbf{u}\|_2 - \mu_r)^2, \quad \mathbf{u} \in \mathbb{R}^5. \quad (17)$$

The model is best understood in polar coordinates: the angular components are uniformly distributed and the radial direction is Gaussian with mean $\mu_r$ and precision $\gamma_r$. The mean is assumed known, and the task is to estimate the precision parameter $\gamma_r$. Figure 1 shows the (normalised) pdf for the ring model in two dimensions, as well as the NCE noise and the CNCE noise generated according to Equation (13). As often done in NCE, a Gaussian noise is chosen to match the mean and covariance of the data distribution. Because of the manifold structure of the data, the NCE noise is concentrated in areas where the data distribution takes small values, which is in contrast to the CNCE noise that well covers the data manifold.

## 3.2. Results

Figures 2a and 2b show the estimation error as a function of the number of data points $N$. For both the Gaussian and ICA models, the CNCE error decreases linearly in the log-log domain as the sample size increases, which indicates convergence in quadratic mean, and hence consistency. Furthermore, as the number of noise-per-data points $\kappa$ grows, the error appears to approach the MLE error.

The MLE of the ICA model had a tendency to get stuck in local minima for a small part of the estimations (13 out of 100). Consequently, the 0.9 quantile for MLE in Figure 2b shows a high and relatively constant error corresponding to such local minima. While this also occurred for CNCE, it is not visible in Figure 2b as it occurred less often (7/100 simulations).

As shown in Figure 2c, NCE performs better than CNCE for the Gaussian model given the same number of noise and data samples. For the ICA model, they are roughly on-par for sufficiently many data samples, see Figure 2d. An advantage for NCE on these models may not be surprising given that the NCE noise distribution already covers the data distribution very well. Furthermore, Figures 2e and 2f show that the difference between NCE and CNCE decreases as ratio of noise to data samples increases.

Figure 3 shows the results for the ring model using $\kappa = 10$. CNCE achieves about one order of magnitude lower estimation error compared to NCE. With reference to Figure 1, this vast improvement over NCE can be understood as follows: For the noise distribution used in NCE, the majority of the noise samples end up inside the ring where the data sample probability is low, so that they are not useful for learning (the classification problem is too easy, with the noise not providing enough contrast). CNCE, on the other hand, automatically generates suitably contrastive noise on (or close to) the data manifold, which facilitates learning.

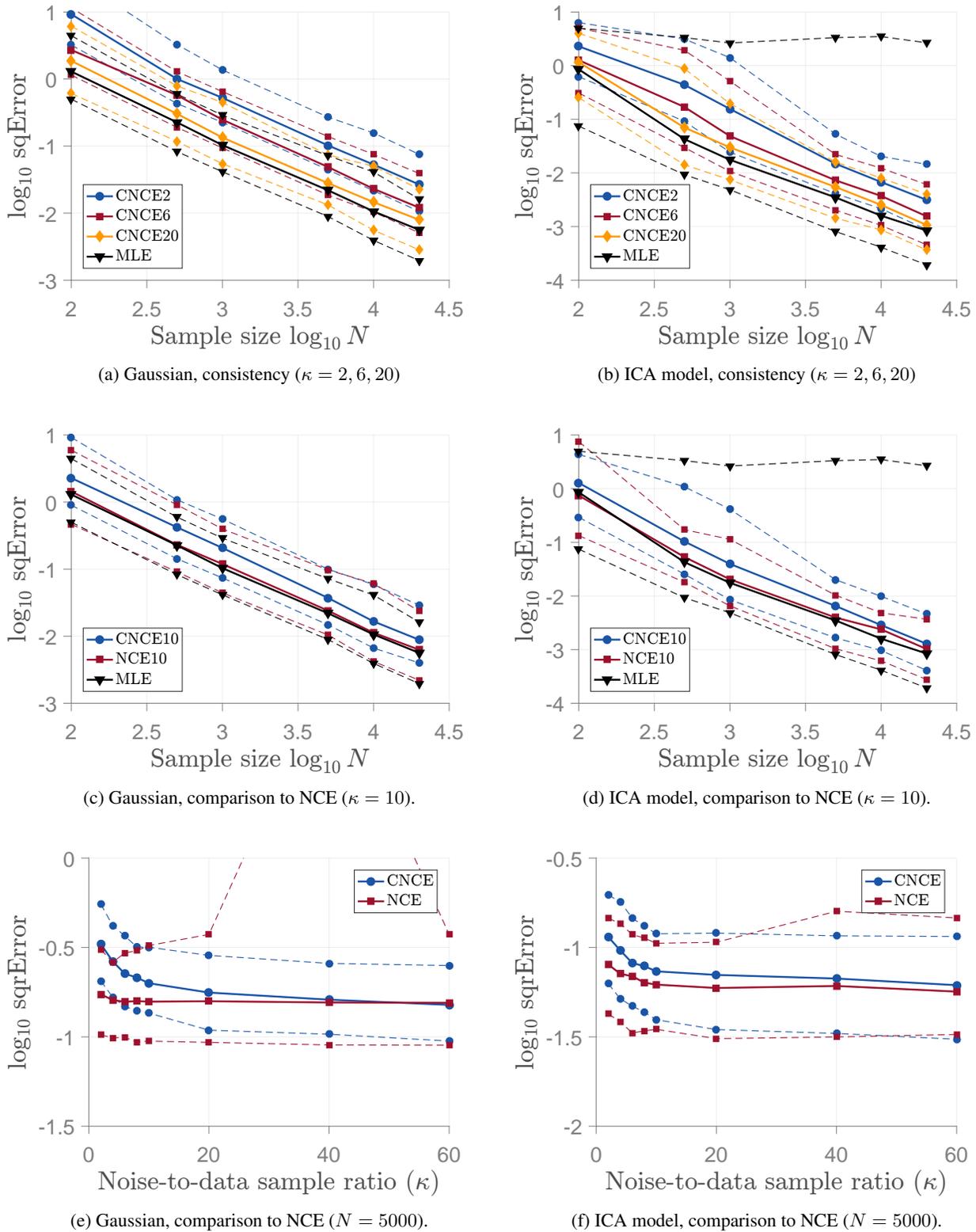

Figure 2: (a-b) CNCE consistency results. (c-d) Comparison to NCE for fixed noise-per-data ratio $\kappa$. (e-f) Comparison for fixed sample size $N$. The solid lines show the median result across 100 different simulations, and the dashed lines the 0.1 and 0.9 quantiles. For each of the 100 simulations, a new random set of data generating parameters was used.

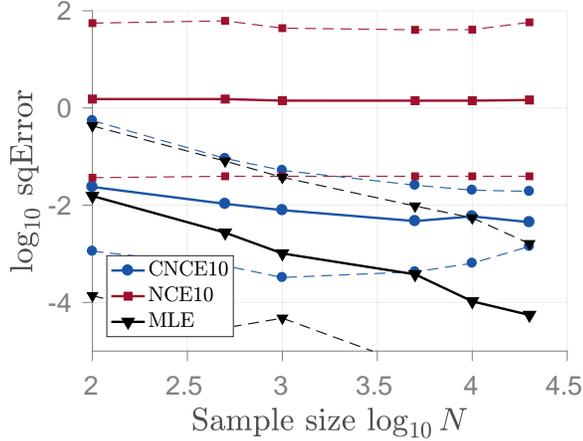

Figure 3: Ring model in 5D, comparison to NCE ($\kappa = 10$)

## 4. Neural image model

To show that CNCE can be used to estimate complex unnormalised models, we used it for unsupervised deep learning and estimated a four-layer feed-forward neural network model from natural images. The model extends the two- and three-layer models of natural images previously estimated with NCE (Gutmann & Hyvärinen, 2012; 2013). We here focus on the learned features. In Supplementary Materials D, we present a qualitative comparison with NCE.

The data $\mathbf{X}$ are image patches of size $32 \times 32$ px, sampled from 11 different monochrome images depicting wild life scenes (van Hateren & van der Schaaf, 1998) in the same manner as (Gutmann & Hyvärinen, 2013). Figure 4a shows examples of the extracted image patches. The sampled image patches were vectorised and both the ensemble mean and local mean (DC component) were subtracted. The resulting data were then whitened and their dimensionality reduced to $D = 600$ by principal component analysis (Murphy, 2012, Chapter 12.2), retaining 98% of the variance. We denote the data (random vector) after preprocessing by $\mathbf{u}^{(1)}$.

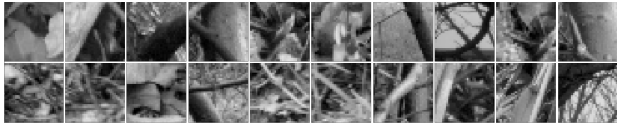

(a) Example of $32 \times 32$ natural image patches.

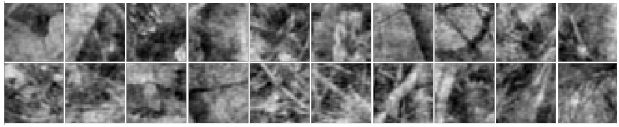

(b) Corresponding noise samples ($\varepsilon = 0.75$).

Figure 4: Data for estimating the deep neural image model.

### 4.1. Model specification

The unnormalised image model $\phi$ defined below consist of a "structured" part $\tilde{\phi}$ that models the non-Gaussianity of the natural image data and a Gaussian part that accounts for the covariance structure. In the PCA space, the model is

$$\log \phi(\mathbf{u}^{(1)}; \boldsymbol{\theta}) = \log \tilde{\phi}(\mathbf{u}^{(1)}; \boldsymbol{\theta}) - \frac{1}{2} \mathbf{u}^{(1)} \cdot \mathbf{u}^{(1)}, \quad (18)$$

where $\cdot$ denotes the inner product between two vectors. This corresponds to a model for images defined in the subspace spanned by the first $D$ principle component directions.

The Gaussian term in (18) tends to mask the non-Gaussian structure that we are primarily interested in. In order to better learn about the non-Gaussian properties of natural images, we define the conditional noise distribution as

$$\log p_c(\mathbf{u}_2|\mathbf{u}_1) = \log \tilde{p}_c(\mathbf{u}_2|\mathbf{u}_1) - \frac{1}{2} \mathbf{u}_2 \cdot \mathbf{u}_2 + \text{const}, \quad (19)$$

where $\tilde{p}_c$ is the Gaussian noise distribution in (13). With this choice, the two Gaussian terms of the model and noise cancel in the nonlinearity $G(\mathbf{u}_1, \mathbf{u}_2; \boldsymbol{\theta})$, so that

$$G(\mathbf{u}_1, \mathbf{u}_2; \boldsymbol{\theta}) = \log \frac{\tilde{\phi}(\mathbf{u}_1; \boldsymbol{\theta}) \tilde{p}_c(\mathbf{u}_2|\mathbf{u}_1)}{\tilde{\phi}(\mathbf{u}_2; \boldsymbol{\theta}) \tilde{p}_c(\mathbf{u}_1|\mathbf{u}_2)}. \quad (20)$$

Due to the cancelling, $\tilde{\phi}$ in Equation (18) is considered the effective model and $\tilde{p}_c$ the effective conditional noise distribution. Examples of noise patches sampled from $\tilde{p}_c$ are shown in Figure 4b.

We next define the (effective) model $\tilde{\phi}$ via a four layers deep, fully connected, feed-forward neural network. The general idea is that we iterate between feature extraction and pooling layers (Gutmann & Hyvärinen, 2013). Unlike in many image models, we here do not impose translation invariance by using convolutional networks; neither do we fix the pooling layers but learn them from data. The input and output dimensions of each layer are provided in Supplementary Materials D.

The preprocessed image patches $\mathbf{u}^{(1)}$ are first passed through a gain-control stage where they are centred and rescaled to cancel out some effects of the lighting conditions (Gutmann & Hyvärinen, 2012),

$$\tilde{\mathbf{u}}(\mathbf{u}) = \sqrt{D-1} \frac{\mathbf{u} - \langle \mathbf{u} \rangle}{\|\mathbf{u} - \langle \mathbf{u} \rangle\|_2}, \quad \langle \mathbf{u} \rangle = \frac{1}{D} \sum_{k=1}^{D} u_k. \quad (21)$$

Then they are passed through a feature extraction and a pooling layer,

$$z_j^{(1)} = \mathbf{w}_j^{(1)} \cdot \tilde{\mathbf{u}}(\mathbf{u}^{(1)}), \quad (22)$$

$$z_j^{(2)} = \log \left( \mathbf{q}_j^{(2)} \cdot \left( \mathbf{z}^{(1)} \right)^2 + 1 \right). \quad (23)$$

Both the features $\mathbf{w}_j^{(1)}$ and pooling weights $\mathbf{q}_j^{(2)}$ are free parameters; we thus learn which 1st layer outputs to pool together. The pooling weights are restricted to be non-negative, which we enforce by writing them as $\mathbf{q}_j^{(2)} = (\mathbf{w}_j^{(2)})^2$, with element-wise squaring. The $\log$ nonlinearity counteracts the squaring, leading to an approximation of the max operation (Gutmann & Hyvärinen, 2013).

We then repeat this processing block of gain control, feature extraction, and pooling: The outputs $z_j^{(2)}$ of the 2nd layer are passed through the same gain control stage as the image patches, i.e. whitening, dimensionality reduction and rescaling, in line with previous work (Gutmann & Hyvärinen, 2013), followed by feature extraction and pooling,

$$z_j^{(3)} = \mathbf{w}_j^{(3)} \cdot \tilde{\mathbf{u}}^{(3)}, \qquad z_j^{(4)} = \mathbf{q}_j^{(4)} \cdot \mathbf{z}^{(3)}. \quad (24)$$

The pooling weights $\mathbf{q}_j^{(4)}$ are restricted to be non-negative, which is enforced as for the second layer. We here work with a simpler pooling model than in Equation (23). An output $z_j^{(4)}$ of the pooling layer is large if $\mathbf{q}_j^{(4)}$ pools over units that are concurrently active, which is related to detecting sign congruency (Gutmann & Hyvärinen, 2009).

The unnormalised model $\tilde{\phi}$ is then given by the total activation of the units in each layer, which means that the overall population activity indicates how likely an input is. Following (Gutmann & Hyvärinen, 2012; 2013) we used

$$\log \tilde{\phi}^{(L)}(\mathbf{u}^{(1)}; \boldsymbol{\theta}) = \sum_{j=1}^{K^{(L)}} f_{\text{th}}\left(z_j^{(L)} + b_j^{(L)}\right) \quad (25)$$

for $L = 2, 3, 4$ where $f_{\text{th}}$ is a smooth rectifying linear unit[1] and $b_j^{(L)}$ threshold parameters that are also learned from the data. The thresholding causes only strongly active units to contribute to $\log \tilde{\phi}^{(L)}(\mathbf{u}^{(1)}; \boldsymbol{\theta})$, which is related to sparse coding (Gutmann & Hyvärinen, 2012). In the case $L = 1$, the outputs $z_j^{(1)}$ were passed through the additional nonlinearity $\log((\cdot)^2 + 1)$ prior to thresholding. This corresponds to computing the 2nd layer outputs with the 2nd layer weights fixed to correspond together to the identity matrix.

We learned the weights hierarchically one layer at a time, e.g. after learning of the 1st layer weights, we kept them fixed and learned the second layer weight vector $\mathbf{w}_j^{(2)}$ etc.

### 4.2. Estimation results

The learned features, i.e receptive fields (RFs) of the 1st layer neurons, can be visualised as images. The learned 2nd layer weight vectors are sparse and the non-zero weights indicate over which 1st layer units the pooling happens. In Figure 5, we visualise randomly selected 2nd layer units, and the 1st layer units that they pool together. The 1st layer has learned Gabor features (Hyvärinen et al., 2009, Chapter 3) and the 2nd layer tends to pool these features according to frequency, orientation and locality, in line with previous models of natural images (Hyvärinen et al., 2009).

To visualise the learned weights on the 3rd layer, we followed (Gutmann & Hyvärinen, 2013) and visualised them as space-orientation receptive fields. That is, we probed the learned neural network with Gabor stimuli at different locations, orientations, and frequency, and visualised the response of the 3rd layer units as a polar plot. The polar plot is centred on the probing location, and the maximal radius is an indicator of the envelope and here spatial frequency of the Gabor stimulus (larger circles correspond to lower spatial frequencies). We visualised the pooling on the 4th layer as for the 2nd layer by indicating the pooling strength with bars underneath the space-orientation receptive fields.

Figure 6 shows examples of the learned 3rd and 4th layer units as well natural image inputs that elicit strong responses for the 4th layer units shown. The learned 3rd layer units detect longer straight or bended contours, which is largely in line with previous findings (Gutmann & Hyvärinen, 2013). The learned 4th layer unit on the top in the figure (unit 4) has learned to pool together 3rd layer units that share the same spatial orientation preference but are tuned to different spatial frequencies. This is line with previous modelling results (Hyvärinen et al., 2005) where similar pooling emerged in a model with more restrictive assumptions. The learned 4th layer unit shown on the bottom (unit 19) is tuned to vertical and horizontal low-frequency structure that bend around the southwest corner, which corresponds to a low-frequency corner detector. The full set of learned units is shown in the same way in Supplementary Materials D. Overall, the results show that CNCE both yields results that are in line with previous work and further finds novel and intuitively reasonable pooling patterns on the newly considered fourth layer.

## 5. Conclusions

In this paper, we addressed the problem of density estimation for unnormalised models where the normalising partition function cannot be computed. We proposed a new method that follows the principles of noise-contrastive estimation and "learning by comparison". In contrast to noise-contrastive estimation (NCE), in the proposed conditional noise-contrastive estimation (CNCE), the contrastive noise is allowed to depend on the data.

The main advantage of allowing the noise distribution to depend on the data is that the information in the data can be leveraged to produce, with rather simple conditional noise distributions as for example a Gaussian, noise samples that are well adapted to a wide range of different data and model

---

[1] $f_{\text{th}}(u) = 0.25 \log(\cosh(2u)) + 0.5u + 0.17$

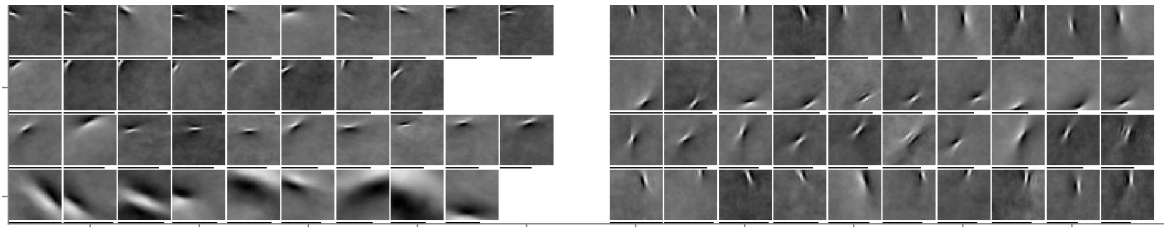

Figure 5: Learned features and pooling for the first two layers of the neural image model. The result for eight units in the $2^{nd}$ layer are shown (each row shows two units). Each icon visualises a $1^{st}$ layer feature, and the thin bar beneath each icon indicates $q_{jk}^{(2)} / \max_k q_{jk}^{(2)}$. Each unit is restricted to show a maximum of ten receptive fields, or as many as to account for 90% of the sum of the second layer weight vector.

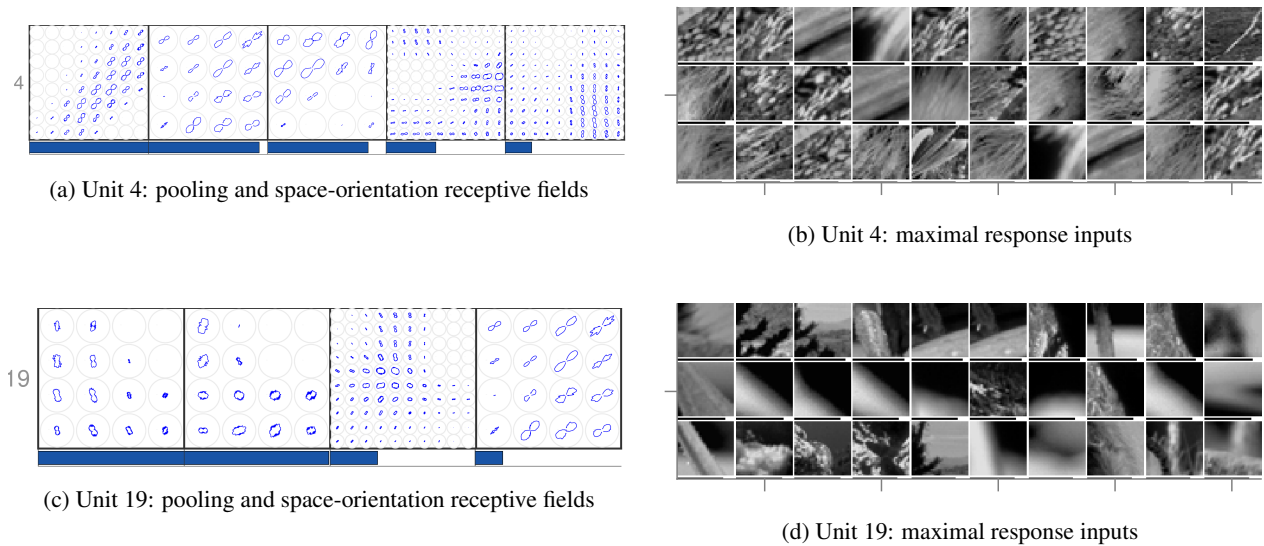

(a) Unit 4: pooling and space-orientation receptive fields

(b) Unit 4: maximal response inputs

(c) Unit 19: pooling and space-orientation receptive fields

(d) Unit 19: maximal response inputs

Figure 6: Examples of learned $3^{rd}$ and $4^{th}$ layer features. The icons show the space-orientation receptive fields, and the bars the pooling strength as in the visualisation of the $2^{nd}$ layer.

types. A second advantage is that for symmetric conditional noise distributions, a closed form expression for the conditional noise is not needed, which both enables a wider choice of distributions and has computational benefits. If the value of the normalisation constant is not of interest, a third advantage of the proposed approach is that the intractable partition function cancels out. Unlike in noise-contrastive estimation, there is thus never a need to introduce an additional parameter for the scaling of the model.

We provided theoretical and empirical arguments that CNCE provides a consistent estimator and proved that score matching emerges as a limiting case. As score matching makes more stringent assumptions but does not rely on sampling, it is an open question whether we can use this result to e.g. devise a hybrid approach where parts of the model are automatically estimated with the more suitable method.

We further found that the relative performances of NCE and CNCE are model dependent, but that CNCE has an advantage in the important case where the data lie in a lower dimensional manifold.

An inherent limitation of empirical comparisons, and hence also those performed here, is that the results depend on the models and noise distributions used. However, given the adaptive nature of CNCE, simple Gaussian conditional noise distributions are likely widely useful, as exemplified by our results on unsupervised deep learning of a neural image model.

The proposed method further allows one to iteratively adapt the conditional noise distribution to make the classification task successively more challenging, as it was done in some simulations for NCE (Gutmann & Hyvärinen, 2010), and generally for learning in generative latent variable models (Gutmann et al., 2014; Goodfellow et al., 2014). This is an interesting direction of future work on CNCE.

## Acknowledgements

MUG would like to thank Jun-ichiro Hirayama at ATR and RIKEN AIP, Japan, for helpful discussions. We thank the anonymous reviewers for their insightful comments.

# Supplementary Material for Conditional Noise-Contrastive Estimation of Unnormalised Models

**Ciwan Ceylan**
UMIC, RWTH Aachen University*
Aachen, Germany
`ceylan@vision.rwth-aachen.de`

**Michael Gutmann**
School of Informatics, University of Edinburgh
Edinburgh, United Kingdom
`michael.gutmann@ed.ac.uk`

## A  Proof of nonparametric estimation theorem

We here prove the consistency theorem for nonparametric estimation. Additionally, an extension to the theorem where the condition $\mathbb{X} = \mathbb{Y}$ is relaxed to $\mathbb{X} \subseteq \mathbb{Y}$ is presented and proved. For this extended theorem, the following definition is required:

$$p_d^{ext}(\mathbf{u}) = \begin{cases} p_d(\mathbf{u}) & \text{if } \mathbf{u} \in \mathbb{X} \\ 0 & \text{if } \mathbf{u} \in \mathbb{Y} \setminus \mathbb{X}. \end{cases} \tag{1}$$

In order to simplify the notation for the proof, the following definitions are introduced,

$$r(\mathbf{u}_1, \mathbf{u}_2) = \frac{p_c(\mathbf{u}_2|\mathbf{u}_1)}{p_c(\mathbf{u}_1|\mathbf{u}_2)} = \frac{1}{r(\mathbf{u}_2, \mathbf{u}_1)}, \tag{2}$$

and

$$\Omega = \{(\mathbf{u}_1, \mathbf{u}_2) \in \mathbb{X} \times \mathbb{X} \mid p_d(\mathbf{u}_1) > 0 \wedge p_c(\mathbf{u}_1|\mathbf{u}_2) > 0\}. \tag{3}$$

Furthermore, the following Taylor expansion will be used in the the proof,

$$\begin{aligned} \log(1 + \exp(-(G + \varepsilon q))) = {} & \log(1 + \exp(-G)) \\ & - \varepsilon q \frac{\exp(-G)}{1 + \exp(-G)} \\ & + \frac{\varepsilon^2 q^2}{2} \frac{\exp(-G)}{(1 + \exp(-G))^2} \\ & + \mathcal{O}(\varepsilon^3). \end{aligned} \tag{4}$$

Using these new definitions, the extended theorem reads:

**Theorem (Nonparametric estimation ext.).** *Let $G : \mathbb{U} \times \mathbb{U} \to \mathbb{R}$ be a function of the form*

$$G(\mathbf{u}_1, \mathbf{u}_2) = f(\mathbf{u}_1) - f(\mathbf{u}_2) + \log r(\mathbf{u}_1, \mathbf{u}_2), \tag{5}$$

*where $f$ is a function from $\mathbb{U}$ to $\mathbb{R}$. Under the assumption $\mathbb{X} \subseteq \mathbb{Y}$, $\tilde{\mathcal{J}}$ attains a unique minimum at*

$$G^*(\mathbf{u}_1, \mathbf{u}_2) = \log \frac{p_d^{ext}(\mathbf{u}_1)p_c(\mathbf{u}_2|\mathbf{u}_1)}{p_d^{ext}(\mathbf{u}_2)p_c(\mathbf{u}_1|\mathbf{u}_2)} \tag{6}$$

*for $(\mathbf{u}_1, \mathbf{u}_2) \in \Omega$.*

First, the proof of the theorem of the main article is presented, followed by the extra steps required to prove the extended theorem.

*Proof of **Nonparametric estimation**.* The proof is divided into two parts. First, $G^*$ is proved to be a critical point of $\tilde{\mathcal{J}}$ by showing that the linear term of the Taylor expansion for $\tilde{\mathcal{J}}$ with respect to $G$ is zero for $G^*$. In the second part, we prove that $G^*$ is a minimum and the only extremum by showing that the quadratic part of the Taylor expansion is strictly positive on the set $\tilde{\Omega}$.

The functional $\tilde{\mathcal{J}}[G]$ is expressed as the integral

$$\tilde{\mathcal{J}}[G] = \mathbb{E}_{\mathbf{xy}} \log\left(1 + \exp(-G(\mathbf{x}, \mathbf{y}))\right) \tag{7}$$

$$= \int_{\mathbb{X} \times \mathbb{Y}} \log\left(1 + \exp(-G(\mathbf{x}, \mathbf{y}))\right) p_d(\mathbf{x}) p_c(\mathbf{x}|\mathbf{y}) \mathrm{d}\mathbf{x} \mathrm{d}\mathbf{y}. \tag{8}$$

Inserting Equation (5), we obtain the functional

$$\tilde{\mathcal{J}}_f[f] = \mathbb{E}_{\mathbf{xy}} \log\left(1 + \exp(f(\mathbf{y}) - f(\mathbf{x}) + \log r(\mathbf{x}, \mathbf{y}))\right)$$
$$= \int_{\mathbb{X} \times \mathbb{Y}} \log\left(1 + \exp(f(\mathbf{y}) - f(\mathbf{x}) + \log r(\mathbf{x}, \mathbf{y}))\right) p_d(\mathbf{x}) p_c(\mathbf{x}|\mathbf{y}) \mathrm{d}\mathbf{x} \mathrm{d}\mathbf{y}. \tag{9}$$

Now consider an arbitrary perturbation $\psi : \mathbb{U} \to \mathbb{R}$ of $f$

$$\tilde{\mathcal{J}}_f[f + \varepsilon \psi] = \mathbb{E}_{\mathbf{xy}} \log\left(1 + \exp\left[f(\mathbf{y}) + \varepsilon \psi(\mathbf{y}) - f(\mathbf{x}) - \varepsilon \psi(\mathbf{x}) + \log r(\mathbf{x}, \mathbf{y})\right]\right) \tag{10}$$

$$= \int_{\mathbb{X} \times \mathbb{Y}} \log\left(1 + \exp\left[-(G(\mathbf{x}, \mathbf{y}) + \varepsilon(\psi(\mathbf{x}) - \psi(\mathbf{y})))\right]\right) p_d(\mathbf{x}) p_c(\mathbf{x}|\mathbf{y}) \mathrm{d}\mathbf{x} \mathrm{d}\mathbf{y}. \tag{11}$$

The perturbation of $\tilde{\mathcal{J}}_f[f]$ corresponds to the following perturbation of $\tilde{\mathcal{J}}[G]$,

$$\tilde{\mathcal{J}}[G + \varepsilon(\psi(\mathbf{x}) - \psi(\mathbf{y}))] = \mathbb{E}_{\mathbf{xy}} \log\left(1 + \exp[-(G(\mathbf{x}, \mathbf{y}) + \varepsilon(\psi(\mathbf{x}) - \psi(\mathbf{y})))]\right). \tag{12}$$

Using the Taylor expansion from Equation (4) gives

$$\tilde{\mathcal{J}}[G + \varepsilon(\psi(\mathbf{x}) - \psi(\mathbf{y}))] = \mathbb{E}_{\mathbf{xy}} \log(1 + \exp(-G(\mathbf{x}, \mathbf{y})))$$
$$- \varepsilon \mathbb{E}_{\mathbf{xy}} (\psi(\mathbf{x}) - \psi(\mathbf{y})) \frac{\exp(-G(\mathbf{x}, \mathbf{y}))}{1 + \exp(-G(\mathbf{x}, \mathbf{y}))}$$
$$+ \frac{\varepsilon^2}{2} \mathbb{E}_{\mathbf{xy}} (\psi(\mathbf{x}) - \psi(\mathbf{y}))^2 \frac{\exp(-G(\mathbf{x}, \mathbf{y}))}{(1 + \exp(-G(\mathbf{x}, \mathbf{y})))^2}$$
$$+ \mathcal{O}(\varepsilon^3). \tag{13}$$

Equating the 1st order term with 0 lets us find a necessary condition for the optimal $G$,

$$0 = \mathbb{E}_{\mathbf{xy}} (\psi(\mathbf{x}) - \psi(\mathbf{y})) \frac{\exp(-G(\mathbf{x}, \mathbf{y}))}{1 + \exp(-G(\mathbf{x}, \mathbf{y}))} \tag{14}$$

$$= \int_{\mathbb{X} \times \mathbb{Y}} \psi(\mathbf{x}) \frac{\exp(-G(\mathbf{x}, \mathbf{y}))}{1 + \exp(-G(\mathbf{x}, \mathbf{y}))} p_d(\mathbf{x}) p_c(\mathbf{y}|\mathbf{x}) \mathrm{d}\mathbf{x} \mathrm{d}\mathbf{y}$$
$$- \int_{\mathbb{X} \times \mathbb{Y}} \psi(\mathbf{y}) \frac{\exp(-G(\mathbf{x}, \mathbf{y}))}{1 + \exp(-G(\mathbf{x}, \mathbf{y}))} p_d(\mathbf{x}) p_c(\mathbf{y}|\mathbf{x}) \mathrm{d}\mathbf{x} \mathrm{d}\mathbf{y} \tag{15}$$

We now make a change of variables. For the first term in Equation (15) we write $\mathbf{u}$ for $\mathbf{x}$ and $\mathbf{v}$ for $\mathbf{y}$ while for the second term we use the transform

$$T_2 : \begin{pmatrix} \mathbf{u} \\ \mathbf{v} \end{pmatrix} = \begin{pmatrix} \mathbf{0} & \mathbb{1} \\ \mathbb{1} & \mathbf{0} \end{pmatrix} \begin{pmatrix} \mathbf{x} \\ \mathbf{y} \end{pmatrix} \tag{16}$$

$$\det \begin{pmatrix} \mathbf{0} & \mathbb{1} \\ \mathbb{1} & \mathbf{0} \end{pmatrix} = -1 \tag{17}$$

$$T_2(\mathbb{X} \times \mathbb{Y}) = \mathbb{Y} \times \mathbb{X}. \tag{18}$$

In the resulting equation, the integrals for the two terms are taken over different domains,

$$
0 = \int_{\mathbb{X} \times \mathbb{Y}} \psi(\mathbf{u}) \frac{\exp(-G(\mathbf{u}, \mathbf{v}))}{1 + \exp(-G(\mathbf{u}, \mathbf{v}))} p_d(\mathbf{u}) p_c(\mathbf{v}|\mathbf{u}) \mathrm{d}\mathbf{u} \mathrm{d}\mathbf{v}
$$
$$
- \int_{\mathbb{Y} \times \mathbb{X}} \psi(\mathbf{u}) \frac{\exp(-G(\mathbf{v}, \mathbf{u}))}{1 + \exp(-G(\mathbf{v}, \mathbf{u}))} p_d(\mathbf{v}) p_c(\mathbf{u}|\mathbf{v}) \mathrm{d}\mathbf{u} \mathrm{d}\mathbf{v}. \tag{19}
$$

For the first theorem we assume $\mathbb{Y} = \mathbb{X}$ so that

$$
0 = \int_{\mathbb{X} \times \mathbb{X}} \psi(\mathbf{u}) \frac{\exp(-G(\mathbf{u}, \mathbf{v}))}{1 + \exp(-G(\mathbf{u}, \mathbf{v}))} p_d(\mathbf{u}) p_c(\mathbf{v}|\mathbf{u}) \mathrm{d}\mathbf{u} \mathrm{d}\mathbf{v}
$$
$$
- \int_{\mathbb{X} \times \mathbb{X}} \psi(\mathbf{u}) \frac{\exp(-G(\mathbf{v}, \mathbf{u}))}{1 + \exp(-G(\mathbf{v}, \mathbf{u}))} p_d(\mathbf{v}) p_c(\mathbf{u}|\mathbf{v}) \mathrm{d}\mathbf{u} \mathrm{d}\mathbf{v} \tag{20}
$$
$$
= \int_{\mathbb{X} \times \mathbb{X}} \psi(\mathbf{u}) \left( \frac{\exp(-G(\mathbf{u}, \mathbf{v})) p_d(\mathbf{u}) p_c(\mathbf{v}|\mathbf{u})}{1 + \exp(-G(\mathbf{u}, \mathbf{v}))} \right.
$$
$$
\left. - \frac{\exp(-G(\mathbf{v}, \mathbf{u})) p_d(\mathbf{v}) p_c(\mathbf{u}|\mathbf{v})}{1 + \exp(-G(\mathbf{v}, \mathbf{u}))} \right) \mathrm{d}\mathbf{u} \mathrm{d}\mathbf{v} \tag{21}
$$

Since Equation (21) should hold for any $\psi$ on $\mathbb{X} \times \mathbb{X}$, the factor in the parenthesis must equal 0. The factor can be expanded by inserting the assumed form of $G$, see Equation (5),

$$
\frac{\exp(-G(\mathbf{u}, \mathbf{v})) p_d(\mathbf{u}) p_c(\mathbf{v}|\mathbf{u})}{1 + \exp(-G(\mathbf{u}, \mathbf{v}))} = \frac{\exp(-G(\mathbf{v}, \mathbf{u})) p_d(\mathbf{v}) p_c(\mathbf{u}|\mathbf{v})}{1 + \exp(-G(\mathbf{v}, \mathbf{u}))} \tag{22}
$$

$$
\frac{p_d(\mathbf{u}) p_c(\mathbf{v}|\mathbf{u})}{\exp(G(\mathbf{u}, \mathbf{v})) + 1} = \frac{p_d(\mathbf{v}) p_c(\mathbf{u}|\mathbf{v})}{\exp(G(\mathbf{v}, \mathbf{u})) + 1} \tag{23}
$$

$$
\frac{p_d(\mathbf{u}) p_c(\mathbf{v}|\mathbf{u})}{\exp(f(\mathbf{u}) - f(\mathbf{v})) r(\mathbf{u}, \mathbf{v}) + 1} = \frac{p_d(\mathbf{v}) p_c(\mathbf{u}|\mathbf{v})}{\exp(f(\mathbf{v}) - f(\mathbf{u})) r(\mathbf{v}, \mathbf{u}) + 1} \tag{24}
$$

$$
\frac{\exp(f(\mathbf{v})) p_d(\mathbf{u}) p_c(\mathbf{v}|\mathbf{u})}{\exp(f(\mathbf{u})) r(\mathbf{u}, \mathbf{v}) + \exp(f(\mathbf{v}))} = \frac{\exp(f(\mathbf{u})) p_d(\mathbf{v}) p_c(\mathbf{u}|\mathbf{v})}{\exp(f(\mathbf{v})) r(\mathbf{v}, \mathbf{u}) + \exp(f(\mathbf{u}))} \tag{25}
$$

Using $r(\mathbf{v}, \mathbf{u}) = 1/r(\mathbf{u}, \mathbf{v})$ from Equation (2), a factor can be taken out of the denominator of the r.h.s,

$$
\frac{\exp(f(\mathbf{v})) p_d(\mathbf{u}) p_c(\mathbf{v}|\mathbf{u})}{\exp(f(\mathbf{u})) r(\mathbf{u}, \mathbf{v}) + \exp(f(\mathbf{v}))} = \frac{1}{r(\mathbf{v}, \mathbf{u})} \frac{\exp(f(\mathbf{u})) p_d(\mathbf{v}) p_c(\mathbf{u}|\mathbf{v})}{\exp(f(\mathbf{v})) + \exp(f(\mathbf{u})) r(\mathbf{u}, \mathbf{v})} \tag{26}
$$

$$
\exp(f(\mathbf{v})) p_d(\mathbf{u}) p_c(\mathbf{v}|\mathbf{u}) = \frac{1}{r(\mathbf{v}, \mathbf{u})} \exp(f(\mathbf{u})) p_d(\mathbf{v}) p_c(\mathbf{u}|\mathbf{v}) \tag{27}
$$

$$
\exp(f(\mathbf{v})) p_d(\mathbf{u}) p_c(\mathbf{v}|\mathbf{u}) = \frac{p_c(\mathbf{v}|\mathbf{u})}{p_c(\mathbf{u}|\mathbf{v})} \exp(f(\mathbf{u})) p_d(\mathbf{v}) p_c(\mathbf{u}|\mathbf{v}). \tag{28}
$$

Now consider only the set $\Omega$ where both sides in the above equation are not trivially zero,

$$
\exp(f(\mathbf{v})) p_d(\mathbf{u}) = \exp(f(\mathbf{u})) p_d(\mathbf{v}) \tag{29}
$$

$$
\frac{p_d(\mathbf{u})}{\exp(f(\mathbf{u}))} = \frac{p_d(\mathbf{v})}{\exp(f(\mathbf{v}))} = Z \tag{30}
$$

$$
f^*(\mathbf{u}) = \log p_d(\mathbf{u}) - \log Z \tag{31}
$$

$$
G^*(\mathbf{u}_1, \mathbf{u}_2) = \log p_d(\mathbf{u}_1) - \log p_d(\mathbf{u}_2) + \log r(\mathbf{u}_1, \mathbf{u}_2). \tag{32}
$$

The first part of the proof is now completed as $G^*$ in Equation (32) is a critical point of $\tilde{\mathcal{J}}$.

It is straightforward to show that $G^*$ is minimising $\tilde{\mathcal{J}}$ and is the only extreme point. By considering the second order term of the Taylor expansion in Equation (13),

$$
\mathbb{E}_{\mathbf{x}\mathbf{y}} (\psi(\mathbf{x}) - \psi(\mathbf{y}))^2 \frac{\exp(-G(\mathbf{x}, \mathbf{y}))}{(1 + \exp(-G(\mathbf{x}, \mathbf{y})))^2}, \tag{33}
$$

we observe that it is positive for all non-constant perturbations $\psi$. Since constant perturbations of $f$ does not change $G$, it can be concluded that Equation (32) describes a minimum and the only extreme point on the set $\Omega$. ∎

*Proof of **Nonparametric estimation ext.**.* We can follow the previous proof until Equation (19), just after the change of variables. We now observe the following

$$\mathbb{X} \times \mathbb{Y} = (\mathbb{X} \cap \mathbb{Y}) \times (\mathbb{X} \cap \mathbb{Y}) \cup (\mathbb{X} \setminus \mathbb{Y}) \times (\mathbb{X} \cap \mathbb{Y})$$
$$\cup (\mathbb{X} \cap \mathbb{Y}) \times (\mathbb{Y} \setminus \mathbb{X}) \cup (\mathbb{X} \setminus \mathbb{Y}) \times (\mathbb{Y} \setminus \mathbb{X}) \tag{34}$$

The assumption $\mathbb{X} \subseteq \mathbb{Y}$ implies $(\mathbb{X} \setminus \mathbb{Y}) = \emptyset$ and $(\mathbb{X} \cap \mathbb{Y}) = \mathbb{X}$. Therefore,

$$\mathbb{X} \times \mathbb{Y} = \Big( (\mathbb{X} \cap \mathbb{Y}) \times (\mathbb{X} \cap \mathbb{Y}) \Big) \cup \Big( (\mathbb{X} \cap \mathbb{Y}) \times (\mathbb{Y} \setminus \mathbb{X}) \Big) \tag{35}$$

$$= \Big( \mathbb{X} \times \mathbb{X} \Big) \cup \Big( \mathbb{X} \times (\mathbb{Y} \setminus \mathbb{X}) \Big), \tag{36}$$

and similarly

$$\mathbb{Y} \times \mathbb{X} = \Big( \mathbb{X} \times \mathbb{X} \Big) \cup \Big( (\mathbb{Y} \setminus \mathbb{X}) \times \mathbb{X} \Big). \tag{37}$$

It is now possible to reevaluate Equation (19),

$$
\begin{aligned}
0 = &\int_{\mathbb{X} \times \mathbb{X}} \psi(\mathbf{u}) \frac{\exp(-G(\mathbf{u}, \mathbf{v}))}{1 + \exp(-G(\mathbf{u}, \mathbf{v}))} p_d(\mathbf{u}) p_c(\mathbf{v}|\mathbf{u}) \mathrm{d}\mathbf{u} \mathrm{d}\mathbf{v} \\
&+ \int_{\mathbb{X} \times (\mathbb{Y} \setminus \mathbb{X})} \psi(\mathbf{u}) \frac{\exp(-G(\mathbf{u}, \mathbf{v}))}{1 + \exp(-G(\mathbf{u}, \mathbf{v}))} p_d(\mathbf{u}) p_c(\mathbf{v}|\mathbf{u}) \mathrm{d}\mathbf{u} \mathrm{d}\mathbf{v} \\
&- \int_{\mathbb{X} \times \mathbb{X}} \psi(\mathbf{u}) \frac{\exp(-G(\mathbf{v}, \mathbf{u}))}{1 + \exp(-G(\mathbf{v}, \mathbf{u}))} p_d(\mathbf{v}) p_c(\mathbf{u}|\mathbf{v}) \mathrm{d}\mathbf{u} \mathrm{d}\mathbf{v} \\
&- \int_{(\mathbb{Y} \setminus \mathbb{X}) \times \mathbb{X}} \psi(\mathbf{u}) \frac{\exp(-G(\mathbf{v}, \mathbf{u}))}{1 + \exp(-G(\mathbf{v}, \mathbf{u}))} p_d(\mathbf{v}) p_c(\mathbf{u}|\mathbf{v}) \mathrm{d}\mathbf{u} \mathrm{d}\mathbf{v}
\end{aligned}
\tag{38}
$$

$$
\begin{aligned}
0 = &\int_{\mathbb{X} \times \mathbb{X}} \psi(\mathbf{u}) \left( \frac{\exp(-G(\mathbf{u}, \mathbf{v})) p_d(\mathbf{u}) p_c(\mathbf{v}|\mathbf{u})}{1 + \exp(-G(\mathbf{u}, \mathbf{v}))} \right. \\
&\left. - \frac{\exp(-G(\mathbf{v}, \mathbf{u})) p_d(\mathbf{v}) p_c(\mathbf{u}|\mathbf{v})}{1 + \exp(-G(\mathbf{v}, \mathbf{u}))} \right) \mathrm{d}\mathbf{u} \mathrm{d}\mathbf{v} \\
&+ \int_{\mathbb{X} \times (\mathbb{Y} \setminus \mathbb{X})} \psi(\mathbf{u}) \frac{\exp(-G(\mathbf{u}, \mathbf{v}))}{1 + \exp(-G(\mathbf{u}, \mathbf{v}))} p_d(\mathbf{u}) p_c(\mathbf{v}|\mathbf{u}) \mathrm{d}\mathbf{u} \mathrm{d}\mathbf{v} \\
&- \int_{(\mathbb{Y} \setminus \mathbb{X}) \times \mathbb{X}} \psi(\mathbf{u}) \frac{\exp(-G(\mathbf{v}, \mathbf{u}))}{1 + \exp(-G(\mathbf{v}, \mathbf{u}))} p_d(\mathbf{v}) p_c(\mathbf{u}|\mathbf{v}) \mathrm{d}\mathbf{u} \mathrm{d}\mathbf{v}
\end{aligned}
\tag{39}
$$

Following the previous proof,

$$G(\mathbf{u}_1, \mathbf{u}_2) = \log p_d(\mathbf{u}_1) - \log p_d(\mathbf{u}_2) + \log r(\mathbf{u}_1, \mathbf{u}_2) \tag{40}$$

will set the first term of Equation (39) to 0. By using the expanded data distribution $p_d^{ext}$ from Equation (1) in place of $p_d$, we find

$$G^*(\mathbf{u}_1, \mathbf{u}_2) = \log p_d^{ext}(\mathbf{u}_1) - \log p_d^{ext}(\mathbf{u}_2) + \log r(\mathbf{u}_1, \mathbf{u}_2). \tag{41}$$

Since $G^*$ becomes arbitrarily large on $\mathbb{X} \times (\mathbb{Y} \setminus \mathbb{X})$, the second and third terms of Equation (39) are 0. Again, the second order term is positive for all non-constant perturbations $\psi$ on the set $\Omega$. ∎

## B  Proof of connection to score matching

*Proof of **Connection to score matching**.* We here assume that $\mathbf{y} = \mathbf{x} + \varepsilon \boldsymbol{\xi}$ where $\boldsymbol{\xi}$ is a vector of uncorrelated random variables of mean zero and variance one that are independent from $\mathbf{x}$ and have a symmetric density.

Since $\boldsymbol{\xi}$ has a symmetric density, $p_c$ is symmetric and cancels in the definition of $G(\mathbf{u}_1, \mathbf{u}_2; \boldsymbol{\theta})$,

$$G(\mathbf{u}_1, \mathbf{u}_2; \boldsymbol{\theta}) = \log \frac{\phi(\mathbf{u}_1; \boldsymbol{\theta}) p_c(\mathbf{u}_2|\mathbf{u}_1)}{\phi(\mathbf{u}_2; \boldsymbol{\theta}) p_c(\mathbf{u}_1|\mathbf{u}_2)} = \log \phi(\mathbf{u}_1; \boldsymbol{\theta}) - \log \phi(\mathbf{u}_2; \boldsymbol{\theta}). \tag{42}$$

The loss function thus is

$$\mathcal{J}(\boldsymbol{\theta}) = 2\mathbb{E}_{\mathbf{x}\mathbf{y}} \log\left[1 + \exp\left(-G(\mathbf{x}, \mathbf{y}; \boldsymbol{\theta})\right)\right] \tag{43}$$

$$= 2\mathbb{E}_{\mathbf{x}\mathbf{y}} \log\left[1 + \exp\left(-\log\phi(\mathbf{x}; \boldsymbol{\theta}) + \log\phi(\mathbf{y}; \boldsymbol{\theta})\right)\right]$$

$$= 2\mathbb{E}_{\mathbf{x}\boldsymbol{\xi}} \log\left[1 + \exp\left(-\log\phi(\mathbf{x}; \boldsymbol{\theta}) + \log\phi(\mathbf{x} + \varepsilon\boldsymbol{\xi}; \boldsymbol{\theta})\right)\right] \tag{44}$$

Let us denote the log unnormalised model $\log\phi(\cdot; \boldsymbol{\theta})$ by $f_{\boldsymbol{\theta}}(\cdot)$ so that

$$\mathcal{J}(\boldsymbol{\theta}) = 2\mathbb{E}_{\mathbf{x}\boldsymbol{\xi}} \log\left[1 + \exp\left(-f_{\boldsymbol{\theta}}(\mathbf{x}) + f_{\boldsymbol{\theta}}(\mathbf{x} + \varepsilon\boldsymbol{\xi})\right)\right] \tag{45}$$

By assumption $\varepsilon$ is small so that for any fixed value of $\boldsymbol{\xi}$, we have

$$f_{\boldsymbol{\theta}}(\mathbf{x} + \varepsilon\boldsymbol{\xi}) = f_{\boldsymbol{\theta}}(\mathbf{x}) + \varepsilon\nabla_{\mathbf{x}} f_{\boldsymbol{\theta}}(\mathbf{x})^T\boldsymbol{\xi} + \frac{\varepsilon^2}{2}\boldsymbol{\xi}^T\mathbf{H}_{\boldsymbol{\theta}}(\mathbf{x})\boldsymbol{\xi} + O(\varepsilon^3) \tag{46}$$

where $\mathbf{H}_{\boldsymbol{\theta}}(\mathbf{x})$ is the Hessian with elements $\partial_{x_i}\partial_{x_j} f_{\boldsymbol{\theta}}(\mathbf{x})$. We thus obtain

$$\mathcal{J}(\boldsymbol{\theta}) = 2\mathbb{E}_{\mathbf{x}\boldsymbol{\xi}} \log\left[1 + \exp\left(\varepsilon\nabla_{\mathbf{x}} f_{\boldsymbol{\theta}}(\mathbf{x})^T\boldsymbol{\xi} + \frac{\varepsilon^2}{2}\boldsymbol{\xi}^T\mathbf{H}_{\boldsymbol{\theta}}(\mathbf{x})\boldsymbol{\xi} + O(\varepsilon^3)\right)\right] \tag{47}$$

The function $\log(1 + \exp(v))$ has the following Taylor expansion around $v = 0$,

$$\log(1 + \exp(v)) = \log(2) + \frac{1}{2}v + \frac{1}{8}v^2 + O(v^3), \tag{48}$$

so that

$$\mathcal{J}(\boldsymbol{\theta}) = 2\mathbb{E}_{\mathbf{x}\boldsymbol{\xi}}\left[\log(2) + \frac{1}{2}\varepsilon\nabla_{\mathbf{x}} f_{\boldsymbol{\theta}}(\mathbf{x})^T\boldsymbol{\xi} + \frac{\varepsilon^2}{4}\boldsymbol{\xi}^T\mathbf{H}_{\boldsymbol{\theta}}(\mathbf{x})\boldsymbol{\xi} + O(\varepsilon^3)\right] +$$

$$2\mathbb{E}_{\mathbf{x}\boldsymbol{\xi}}\left[\frac{1}{8}\left(\varepsilon\nabla_{\mathbf{x}} f_{\boldsymbol{\theta}}(\mathbf{x})^T\boldsymbol{\xi} + \frac{\varepsilon^2}{2}\boldsymbol{\xi}^T\mathbf{H}_{\boldsymbol{\theta}}(\mathbf{x})\boldsymbol{\xi} + O(\varepsilon^3)\right)^2\right]. \tag{49}$$

Squaring the term $\left(\varepsilon\nabla_{\mathbf{x}} f_{\boldsymbol{\theta}}(\mathbf{x})^T\boldsymbol{\xi} + \frac{\varepsilon^2}{2}\boldsymbol{\xi}^T\mathbf{H}_{\boldsymbol{\theta}}(\mathbf{x})\boldsymbol{\xi} + O(\varepsilon^3)\right)^2$ gives $\varepsilon^2(\nabla_{\mathbf{x}} f_{\boldsymbol{\theta}}(\mathbf{x})^T\boldsymbol{\xi})^2 + O(\varepsilon^3)$ so that

$$\mathcal{J}(\boldsymbol{\theta}) = 2\mathbb{E}_{\mathbf{x}\boldsymbol{\xi}}\left[\log(2) + \frac{1}{2}\varepsilon\nabla_{\mathbf{x}} f_{\boldsymbol{\theta}}(\mathbf{x})^T\boldsymbol{\xi} + \frac{\varepsilon^2}{4}\boldsymbol{\xi}^T\mathbf{H}_{\boldsymbol{\theta}}(\mathbf{x})\boldsymbol{\xi} + \frac{1}{8}\varepsilon^2(\nabla_{\mathbf{x}} f_{\boldsymbol{\theta}}(\mathbf{x})^T\boldsymbol{\xi})^2\right] + O(\varepsilon^3) \tag{50}$$

By assumption, $\mathbf{x}$ and $\boldsymbol{\xi}$ are independent, and $\mathbb{E}_{\boldsymbol{\xi}}\boldsymbol{\xi} = 0$, so that we have

$$\mathcal{J}(\boldsymbol{\theta}) = 2\mathbb{E}_{\mathbf{x}}\mathbb{E}_{\boldsymbol{\xi}}\left[\log(2) + \frac{\varepsilon^2}{4}\boldsymbol{\xi}^T\mathbf{H}_{\boldsymbol{\theta}}(\mathbf{x})\boldsymbol{\xi} + \frac{1}{8}\varepsilon^2(\nabla_{\mathbf{x}} f_{\boldsymbol{\theta}}(\mathbf{x})^T\boldsymbol{\xi})^2\right]$$

$$+ O(\varepsilon^3) \tag{51}$$

Furthermore, with $\mathbb{E}_{\boldsymbol{\xi}}\boldsymbol{\xi}\boldsymbol{\xi}^T$ being equal to the identity matrix $\mathbb{1}$, we have

$$\mathbb{E}_{\boldsymbol{\xi}}\boldsymbol{\xi}^T\mathbf{H}_{\boldsymbol{\theta}}(\mathbf{x})\boldsymbol{\xi} = \mathbb{E}_{\boldsymbol{\xi}}\operatorname{tr}\left[\mathbf{H}_{\boldsymbol{\theta}}(\mathbf{x})\boldsymbol{\xi}\boldsymbol{\xi}^T\right] \tag{52}$$

$$= \operatorname{tr}\left[\mathbf{H}_{\boldsymbol{\theta}}(\mathbf{x})\mathbb{E}_{\boldsymbol{\xi}}\boldsymbol{\xi}\boldsymbol{\xi}^T\right] \tag{53}$$

$$= \operatorname{tr}\mathbf{H}_{\boldsymbol{\theta}}(\mathbf{x}). \tag{54}$$

Similarly, we obtain

$$\mathbb{E}_{\boldsymbol{\xi}}(\nabla_{\mathbf{x}} f_{\boldsymbol{\theta}}(\mathbf{x})^T\boldsymbol{\xi})^2 = \mathbb{E}_{\boldsymbol{\xi}}\boldsymbol{\xi}^T\nabla_{\mathbf{x}} f_{\boldsymbol{\theta}}(\mathbf{x})\nabla_{\mathbf{x}} f_{\boldsymbol{\theta}}(\mathbf{x})^T\boldsymbol{\xi} \tag{55}$$

$$= \mathbb{E}_{\boldsymbol{\xi}}\operatorname{tr}\left[\nabla_{\mathbf{x}} f_{\boldsymbol{\theta}}(\mathbf{x})\nabla_{\mathbf{x}} f_{\boldsymbol{\theta}}(\mathbf{x})^T\boldsymbol{\xi}\boldsymbol{\xi}^T\right] \tag{56}$$

$$= \operatorname{tr}\left[\nabla_{\mathbf{x}} f_{\boldsymbol{\theta}}(\mathbf{x})\nabla_{\mathbf{x}} f_{\boldsymbol{\theta}}(\mathbf{x})^T\mathbb{E}_{\boldsymbol{\xi}}\boldsymbol{\xi}\boldsymbol{\xi}^T\right] \tag{57}$$

$$= \operatorname{tr}\left[\nabla_{\mathbf{x}} f_{\boldsymbol{\theta}}(\mathbf{x})\nabla_{\mathbf{x}} f_{\boldsymbol{\theta}}(\mathbf{x})^T\right] \tag{58}$$

$$= ||\nabla_{\mathbf{x}} f_{\boldsymbol{\theta}}(\mathbf{x})||_2^2. \tag{59}$$

With both identities plugged into (51), we can write $\mathcal{J}(\boldsymbol{\theta})$ as

$$\mathcal{J}(\boldsymbol{\theta}) = \frac{\varepsilon^2}{2}\mathbb{E}_\mathbf{x}\left[\operatorname{tr}\mathbf{H}_{\boldsymbol{\theta}}(\mathbf{x}) + \frac{1}{2}||\nabla_\mathbf{x}f_{\boldsymbol{\theta}}(\mathbf{x})||_2^2\right] + 2\log(2) + O(\varepsilon^3). \tag{60}$$

Since $\operatorname{tr}\mathbf{H}_{\boldsymbol{\theta}}(\mathbf{x})$ equals the sum of the second derivatives of $f_{\boldsymbol{\theta}}(\mathbf{x}) = \log\phi(\mathbf{x};\boldsymbol{\theta})$,

$$\operatorname{tr}\mathbf{H}_{\boldsymbol{\theta}}(\mathbf{x}) = \sum_i \frac{\partial^2 f_{\boldsymbol{\theta}}(\mathbf{x})}{\partial x_i^2} \tag{61}$$

the term in the brackets is the loss function that is minimised in score matching, which completes the proof. ∎

## C   Empirical validation on non-negative and discrete data

CNCE was also verified on a heavy-tailed distribution of positive data (log-normal) and a discrete distribution (Bernoulli).

***The log-normal distribution*** is a univariate continuous heavy-tailed distribution that is defined to have its samples normal distributed in the log domain. Consequently, it is only defined on the positive real axis $\mathbb{X} = \mathbb{R}^+$. For this reason, the log-normal distribution is suitable to illustrate the fact that only $\mathbb{X} \subseteq \mathbb{Y}$ is required for CNCE given that the conditional noise distribution $p_c$ defined in the main paper generates noise samples in $\mathbb{Y} = \mathbb{R}$. We used the following unnormalised log-normal model defined over the whole real axis

$$\log\phi(u;\theta,C) = \begin{cases} -\frac{\theta}{2}(\log u)^2 - \log u & \text{if } u > 0 \\ C & \text{if } u \le 0 \end{cases} \tag{62}$$

where $\theta, C \in \mathbb{R}$. On the positive axis, the model is proportional to a log-normal distribution with mean zero in the log domain and with precision $\theta$. On the negative axis, the model assumes the constant value $C$. In theory, the optimal value of $C$ would be $-\infty$. Since this can never be reached in practice, we only measured the estimation error for $\theta$ as the absolute error between true and estimated parameter.

***The Bernoulli model*** defines a simple probability mass function for a binary random variable taking values on $\mathbb{X} = \{0, 1\}$. In the normalised version, the Bernoulli model only has one free parameter. Here, an unnormalised version with two free parameters $\theta_1, \theta_2 \in \mathbb{R}^+$ is used,

$$\log\phi(u;\theta_1,\theta_2) = \begin{cases} \log\theta_1 & \text{if } u = 0 \\ \log\theta_2 & \text{if } u = 1. \end{cases} \tag{63}$$

The use of two free parameters means that there exist an infinite set of equivalent model parameters which only differs from $(\theta_1, \theta_2)$ by a scaling factor. Consequently, to measure the error for a parameter estimate of the Bernoulli model, i.e. $\hat{\boldsymbol{\theta}} = (\hat{\theta}_1, \hat{\theta}_2)$, we normalised the parameters before computing the estimation error as $||(\hat{\theta}_1 + \hat{\theta}_2)^{-1}\hat{\boldsymbol{\theta}} - \boldsymbol{\theta}^*||_2$, where $\boldsymbol{\theta}^* = (\theta_1^*, \theta_2^*)$ denotes the true parameter values (which are related by $\theta_2^* = 1 - \theta_1^*$).

The discrete conditional noise distribution defined by Equation (64) was used for the Bernoulli model. Again $\varepsilon$ controls the similarity between data and noise, but with the added restriction $\varepsilon \in [0, 1]$.

$$p_c^{Ber}(y|x;\varepsilon) = \begin{cases} 1 - \varepsilon & \text{if } y = x \\ \varepsilon & \text{if } y \ne x, \end{cases} \tag{64}$$

## D   Supplemental feature visualisations

### D.1   Neural network layer sizes

The sizes of the four layers are provided in Table 1. Note that the dimensionality of the data was reduced by four using PCA as part of the gain control between the $2^{\text{nd}}$ and $3^{\text{rd}}$ layers.

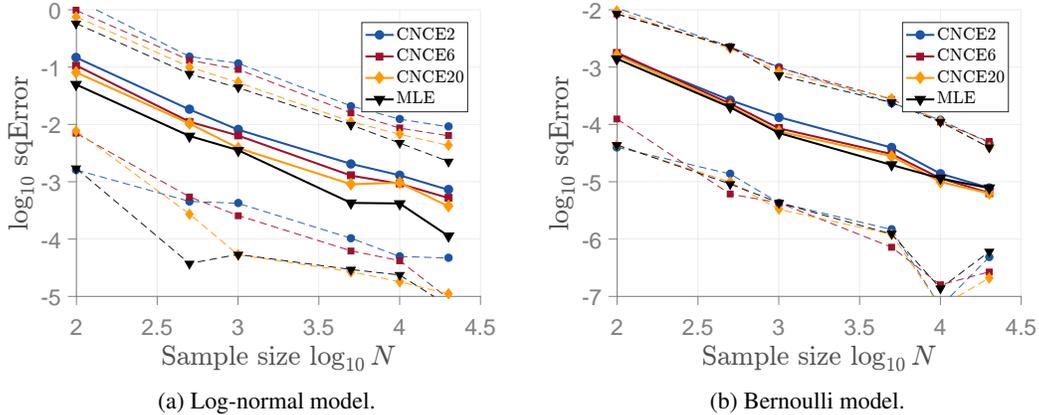

(a) Log-normal model.

(b) Bernoulli model.

Figure 1: Empirical validation of consistency for CNCE. The x-axis shows the sample size in the $\log_{10}$ domain, the y-axis the squared estimation error in the $\log_{10}$ domain. The solid lines show the median result across 100 different simulations, and the dashed lines are the 0.1 and 0.9 quantiles. For each of the 100 simulations, a new random set of parameters were used to generate the data. The different coloured and marked lines correspond to different values of $\kappa$ for CNCE and the black line to the MLE results.

Table 1: Neural network input and output dimensions.

| Layer | Input $D^{(L)}$ | Output $K^{(L)}$ |
|---|---|---|
| 1 | 600 | 600 |
| 2 | 600 | 200 |
| Intermediate gain control | | |
| 3 | 196 | 60 |
| 4 | 60 | 30 |

## D.2 CNCE and NCE 1st layer features comparison

In addition to the quantitative comparisons between CNCE and NCE, which were presented in Section 3 of the main paper, it is desirable to evaluate qualitative differences between the methods. To this end we compared the easily interpretable 1st layer features at different stages of training with the aim to determine qualitative differences between learned features and if learning is faster for one method or the other.

Figure 2 shows the common initialisation and Figures 3 to 13 one hundred 1st layer features at the end of the first eleven meta-iterations. Each meta-iteration consists of ten gradient steps after which new noise samples are generated. The methods seem to learn similar features and for this model, while we do not want to claim superior performance given the qualitative nature of the comparison, CNCE does appear to learn slightly faster than NCE.

## D.3 3rd layer features

All 60 3rd layer space-orientation receptive fields and maximal response patches are shown in Figures 14 to 17.

## D.4 4th layer features

All 30 4th units are visualised in Figures 18 to 47 in the same manner as in the main text.

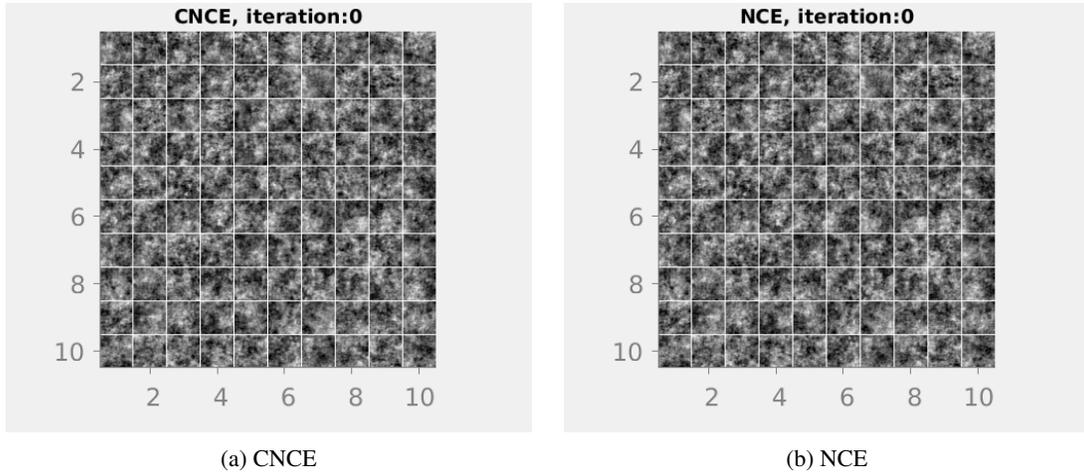

(a) CNCE

(b) NCE

Figure 2: The common initialisation for the 100 1st later features used for the qualitative comparison between CNCE and NCE.

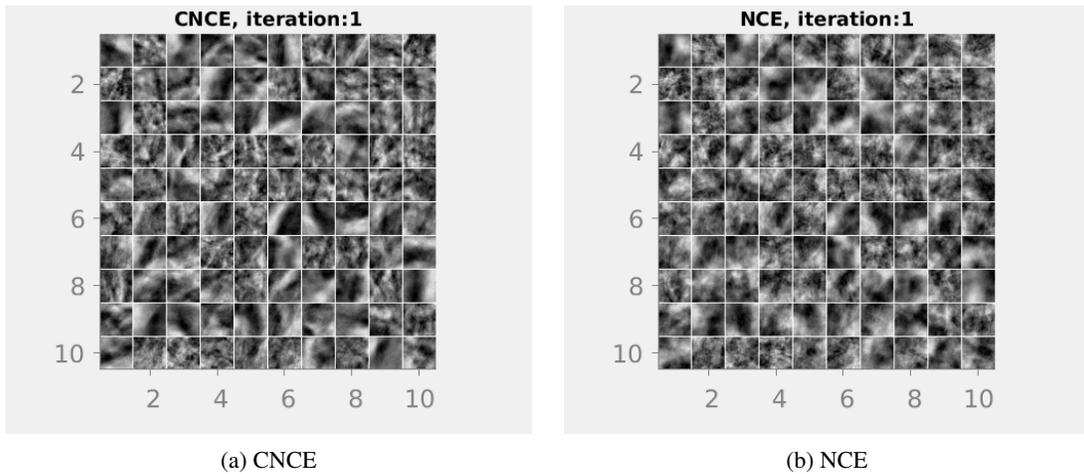

(a) CNCE

(b) NCE

Figure 3: A sample of the 1st layer features after 1 meta-iteration.

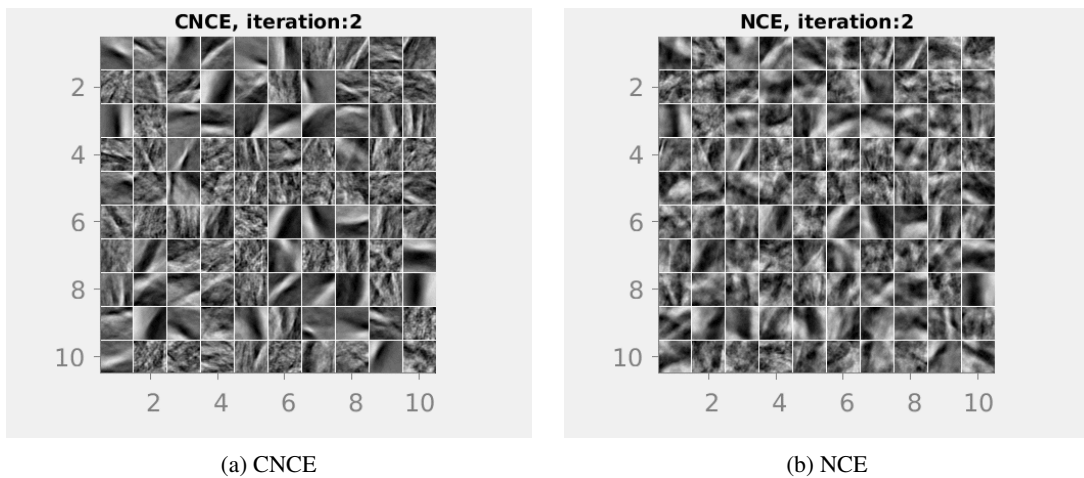

(a) CNCE

(b) NCE

Figure 4: A sample of the 1st layer features after 2 meta-iterations.

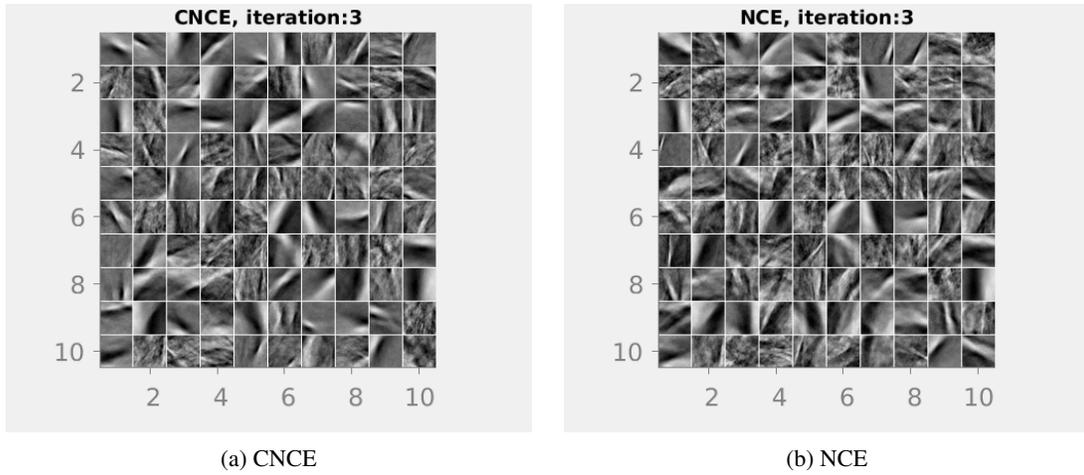

(a) CNCE                                    (b) NCE

Figure 5: A sample of the 1$^{st}$ layer features after 3 meta-iterations.

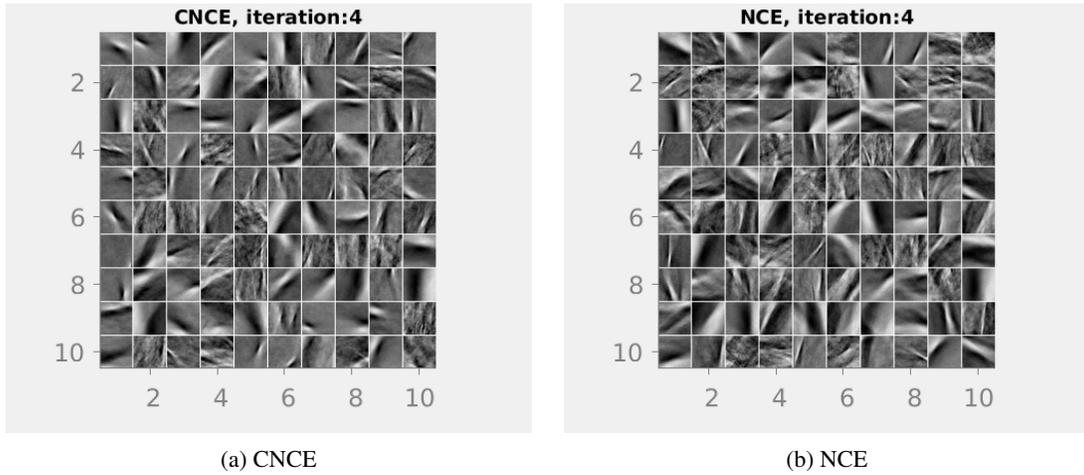

(a) CNCE                                    (b) NCE

Figure 6: A sample of the 1$^{st}$ layer features after 4 meta-iterations.

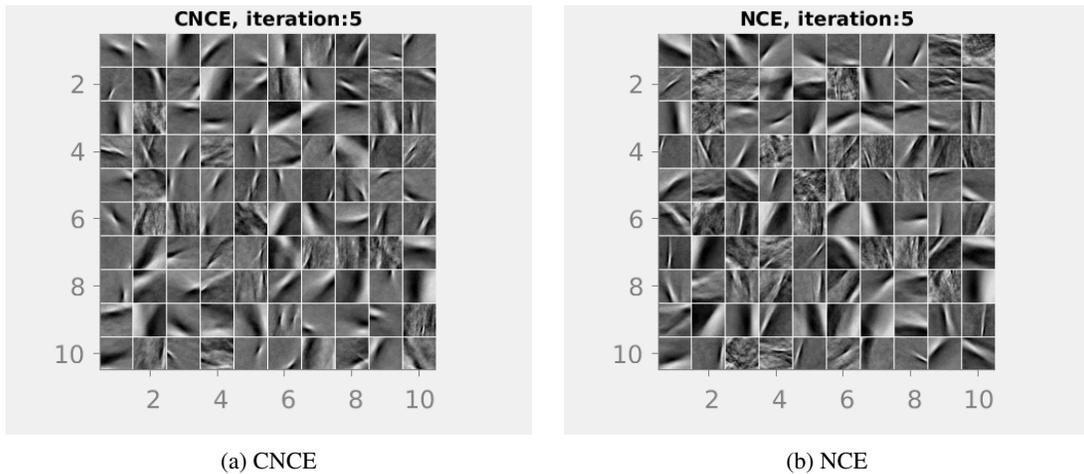

(a) CNCE                                    (b) NCE

Figure 7: A sample of the 1$^{st}$ layer features after 5 meta-iterations.

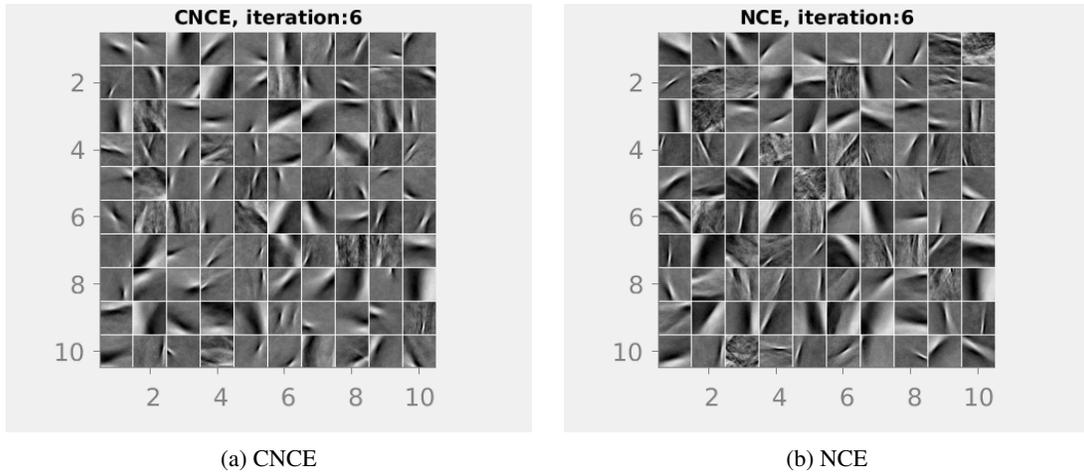

(a) CNCE            (b) NCE

Figure 8: A sample of the 1$^{\text{st}}$ layer features after 6 meta-iterations.

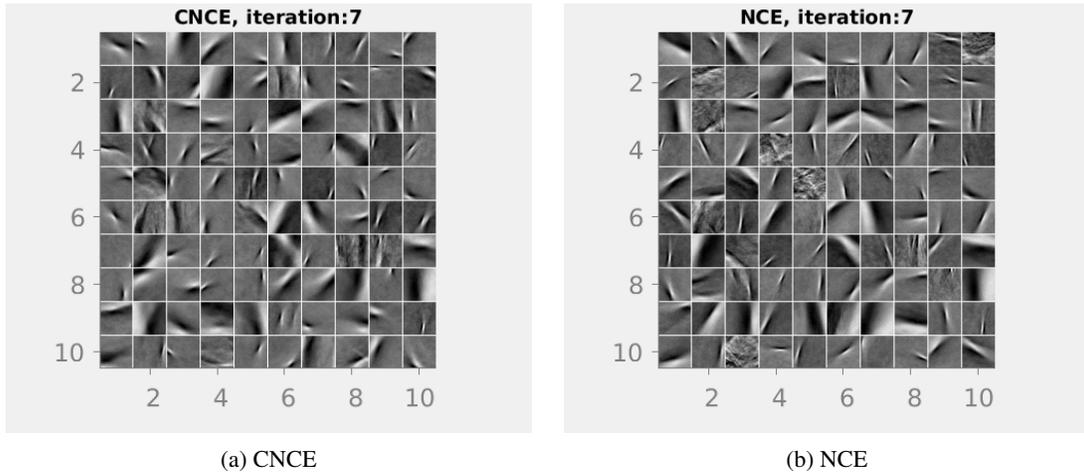

(a) CNCE            (b) NCE

Figure 9: A sample of the 1$^{\text{st}}$ layer features after 7 meta-iterations.

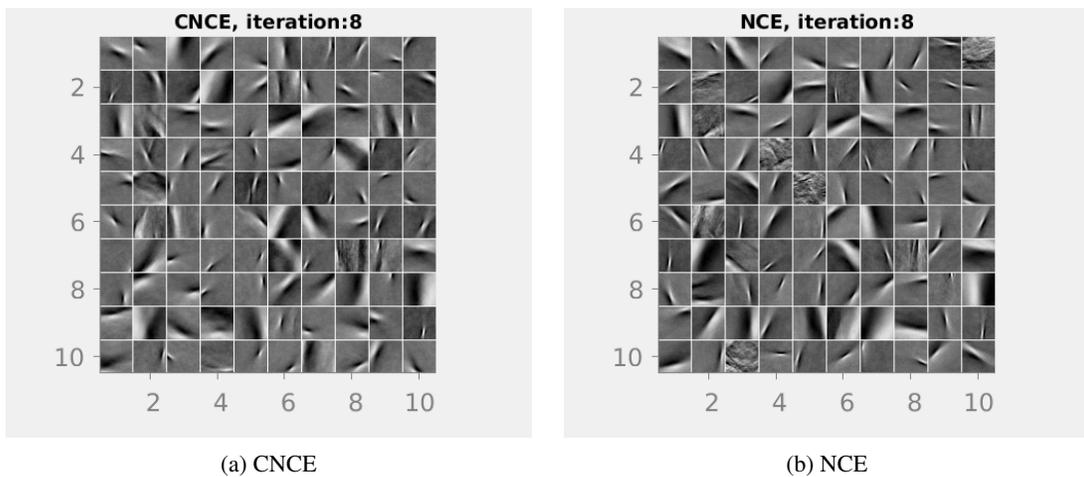

(a) CNCE            (b) NCE

Figure 10: A sample of the 1$^{\text{st}}$ layer features after 8 meta-iterations.

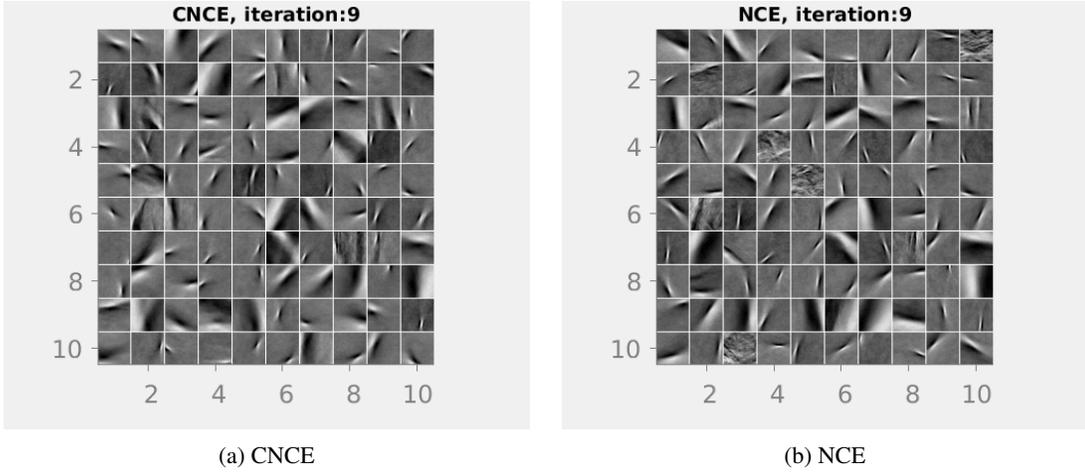

(a) CNCE

(b) NCE

Figure 11: A sample of the 1$^{st}$ layer features after 9 meta-iterations.

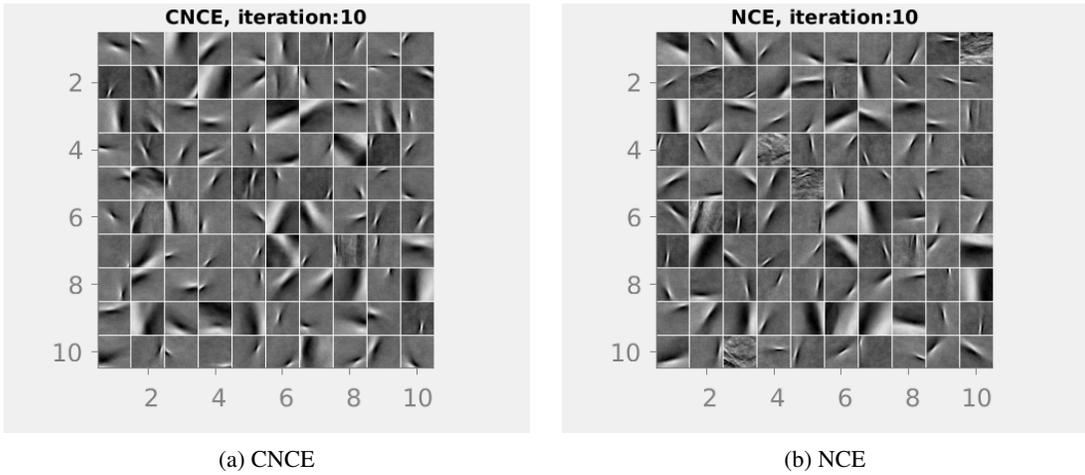

(a) CNCE

(b) NCE

Figure 12: A sample of the 1$^{st}$ layer features after 10 meta-iterations.

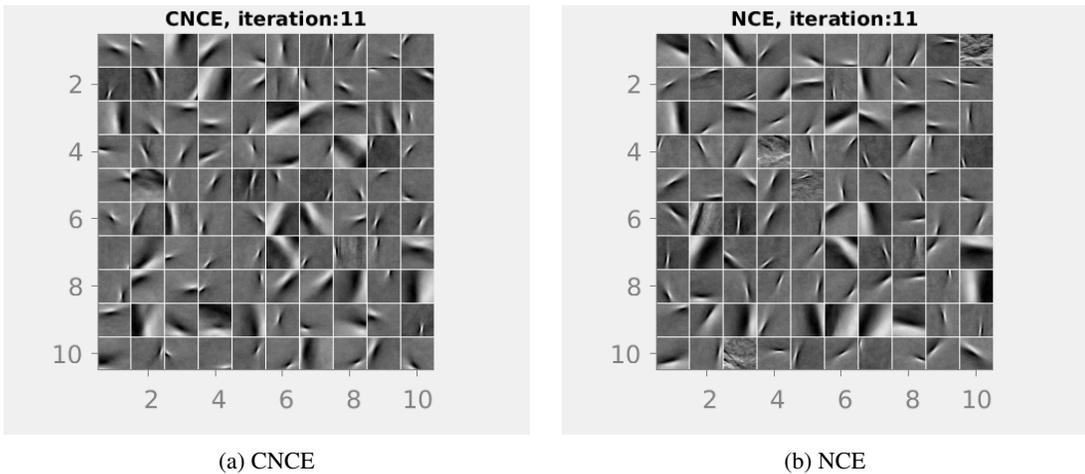

(a) CNCE

(b) NCE

Figure 13: A sample of the 1$^{st}$ layer features after 11 meta-iterations.

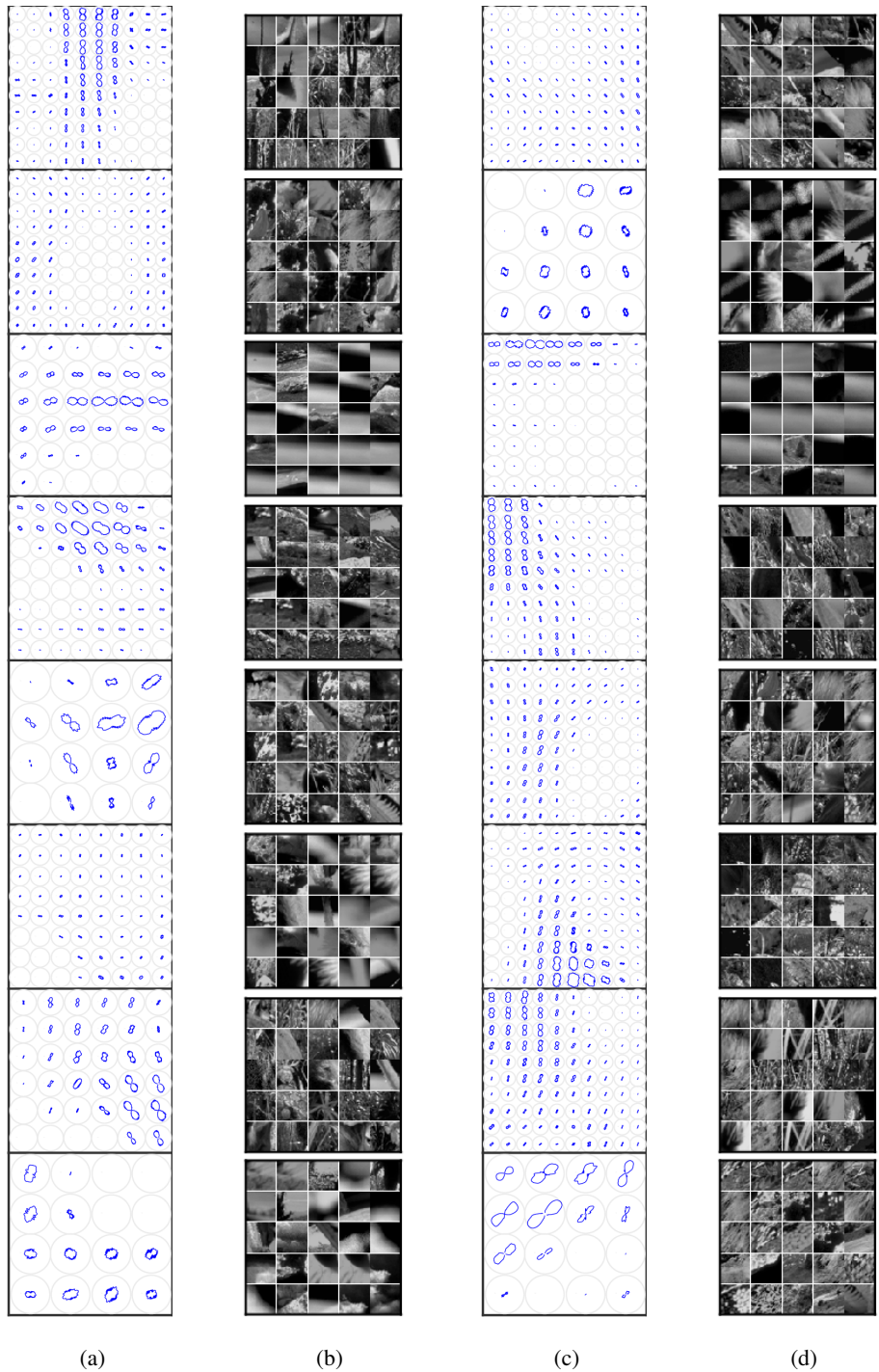

(a)        (b)        (c)        (d)

Figure 14: Each row pair of one receptive field and one icon represent a 3$^{\text{rd}}$ layer unit. (a) and (b) show units 1 to 8, and (c) and (d) 9 to 16.

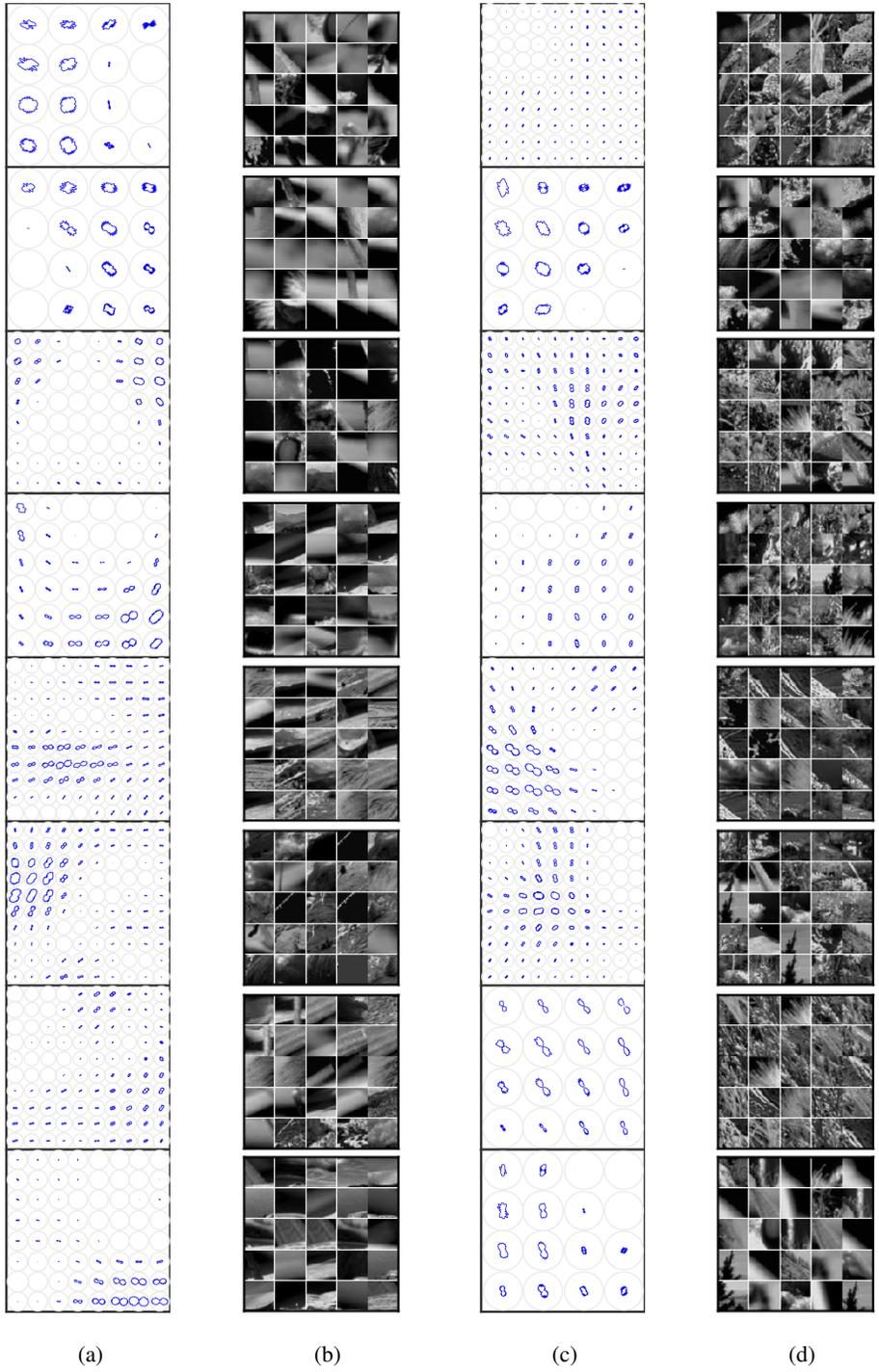

(a)   (b)   (c)   (d)

Figure 15: Each row pair of one receptive field and one icon represent a 3<sup>rd</sup> layer unit. (a) and (b) show units 17 to 24, and (c) and (d) 25 to 32.

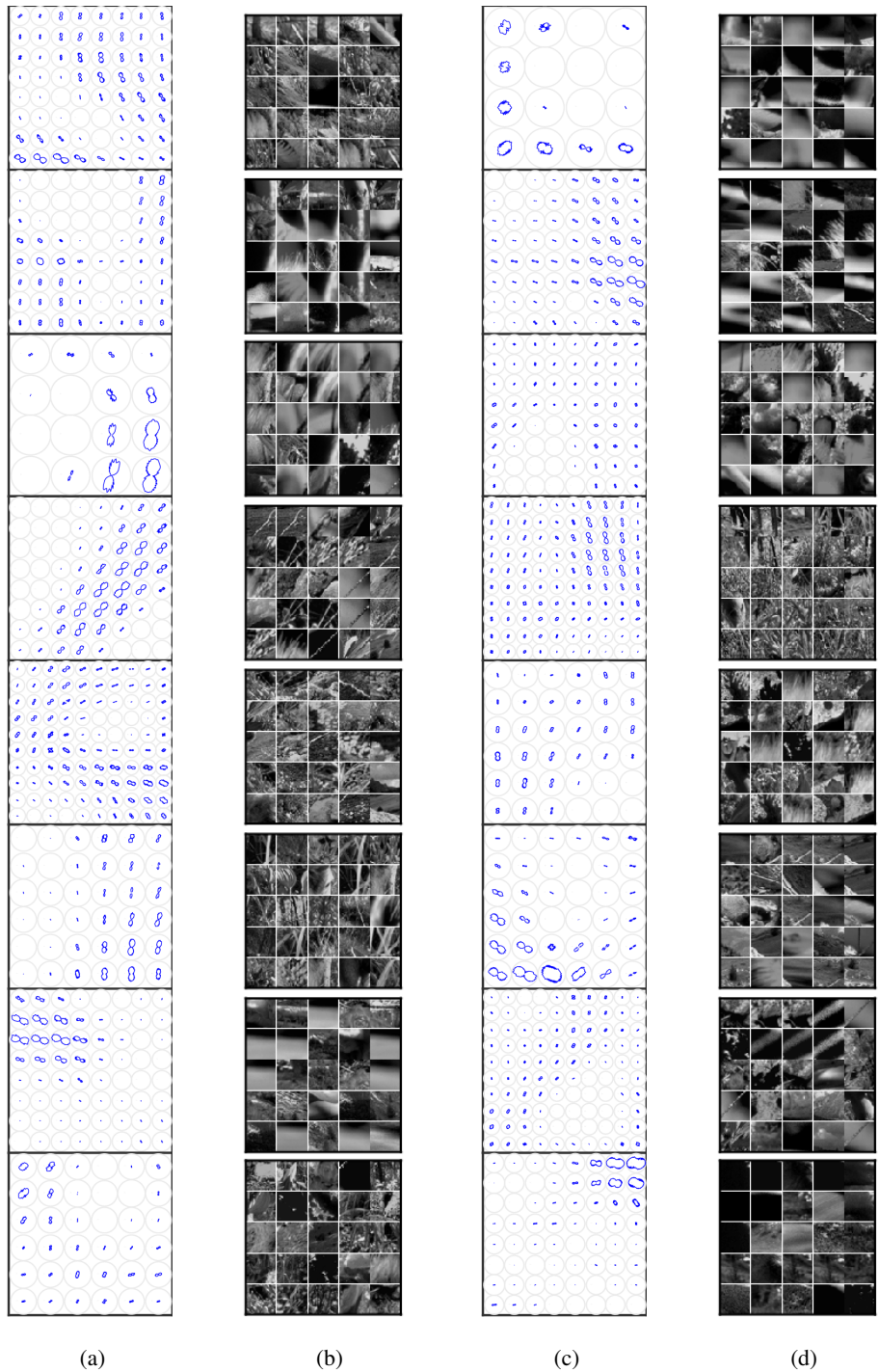

(a)        (b)        (c)        (d)

Figure 16: Each row pair of one receptive field and one icon represent a 3$^{\mathrm{rd}}$ layer unit. (a) and (b) show units 33 to 40, and (c) and (d) 41 to 48.

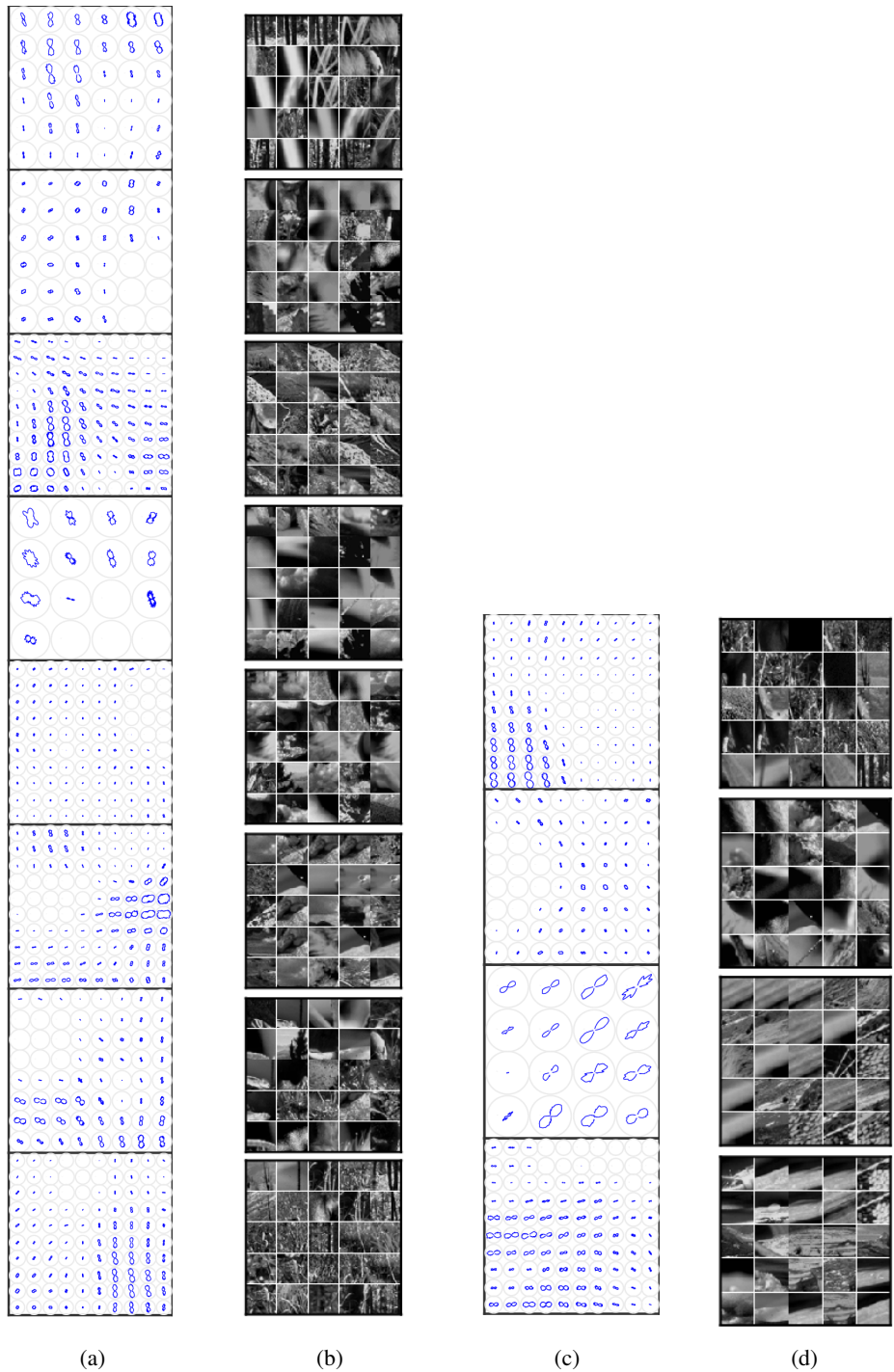

(a)        (b)        (c)        (d)

Figure 17: Each row pair of one receptive field and one icon represent a 3$^{\rm rd}$ layer unit. (a) and (b) show units 49 to 56, and (c) and (d) 57 to 60.

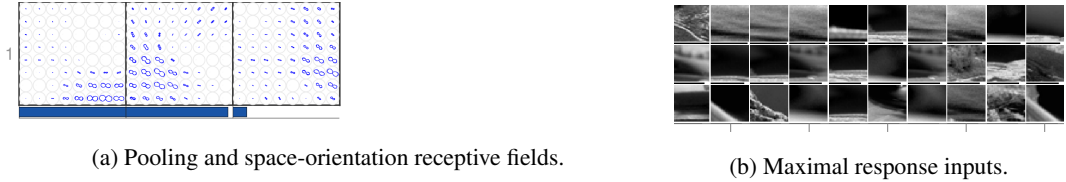

(a) Pooling and space-orientation receptive fields.

(b) Maximal response inputs.

Figure 18: The estimation for unit 1 in the 4th layer. (a) shows the learned pooling of the 3rd layer units and (b) shows 30 image patches that produced maximal responses for a batch of 10000 inputs. For the space-orientation receptive fields, the bar beneath each icon indicates the relative size of the 4th weight, i.e. $q_{1,k}^{(4)} / \max_k q_{1,k}^{(4)}$. The receptive fields shown account for 90% of the sum of the weight vector. The thin bars beneath each image patch indicate the response strength relative to the patch with the maximal response.

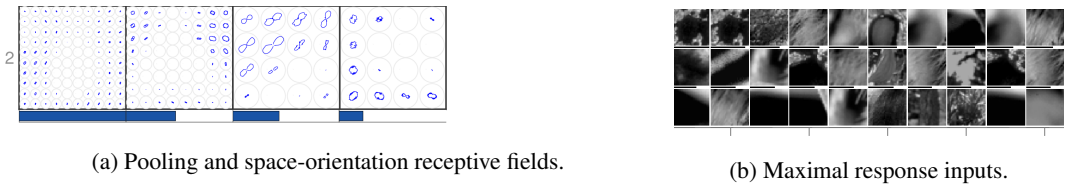

(a) Pooling and space-orientation receptive fields.

(b) Maximal response inputs.

Figure 19: Unit 2 in the 4th layer, visualised as in Figure 18.

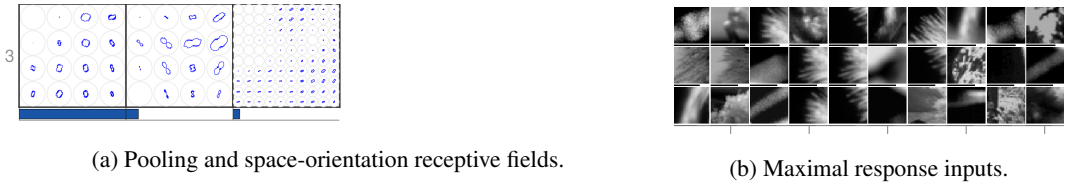

(a) Pooling and space-orientation receptive fields.

(b) Maximal response inputs.

Figure 20: Unit 3 in the 4th layer, visualised as in Figure 18.

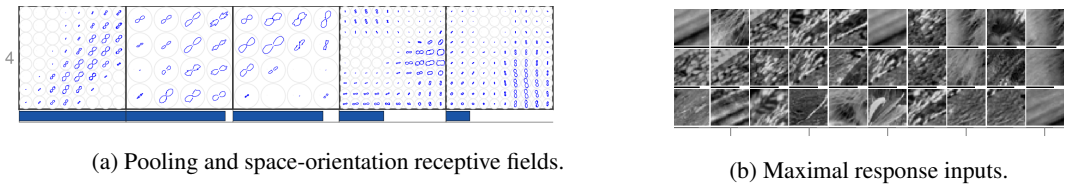

(a) Pooling and space-orientation receptive fields.

(b) Maximal response inputs.

Figure 21: Unit 4 in the 4th layer, visualised as in Figure 18.

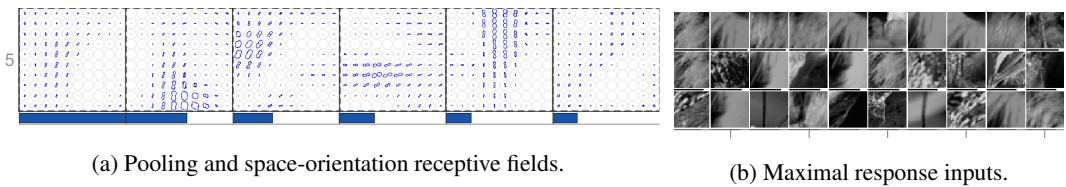

(a) Pooling and space-orientation receptive fields.

(b) Maximal response inputs.

Figure 22: Unit 5 in the 4th layer, visualised as in Figure 18.

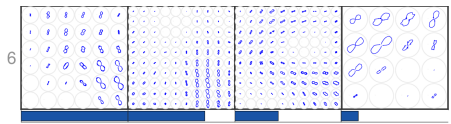

(a) Pooling and space-orientation receptive fields.

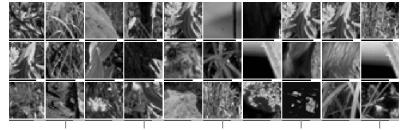

(b) Maximal response inputs.

Figure 23: Unit 6 in the 4$^{th}$ layer, visualised as in Figure 18.

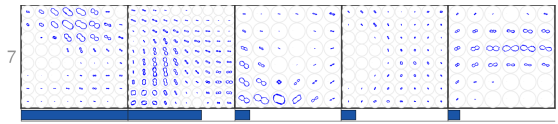

(a) Pooling and space-orientation receptive fields.

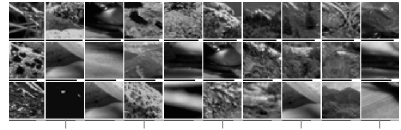

(b) Maximal response inputs.

Figure 24: Unit 7 in the 4$^{th}$ layer, visualised as in Figure 18.

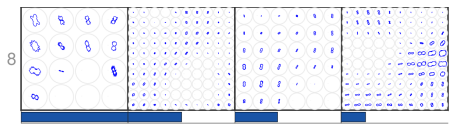

(a) Pooling and space-orientation receptive fields.

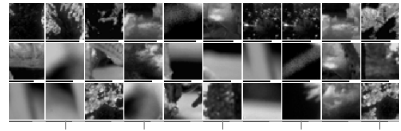

(b) Maximal response inputs.

Figure 25: Unit 8 in the 4$^{th}$ layer, visualised as in Figure 18.

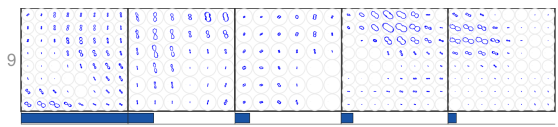

(a) Pooling and space-orientation receptive fields.

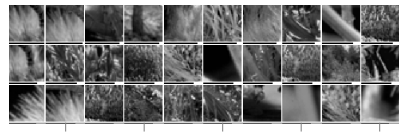

(b) Maximal response inputs.

Figure 26: Unit 9 in the 4$^{th}$ layer, visualised as in Figure 18.

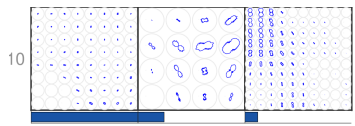

(a) Pooling and space-orientation receptive fields.

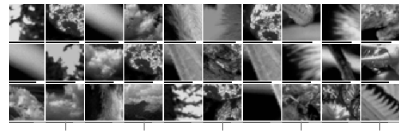

(b) Maximal response inputs.

Figure 27: Unit 10 in the 4$^{th}$ layer, visualised as in Figure 18.

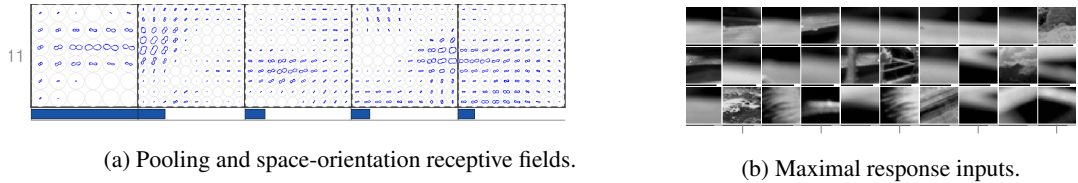

(a) Pooling and space-orientation receptive fields.

(b) Maximal response inputs.

Figure 28: Unit 11 in the 4$^{\text{th}}$ layer, visualised as in Figure 18.

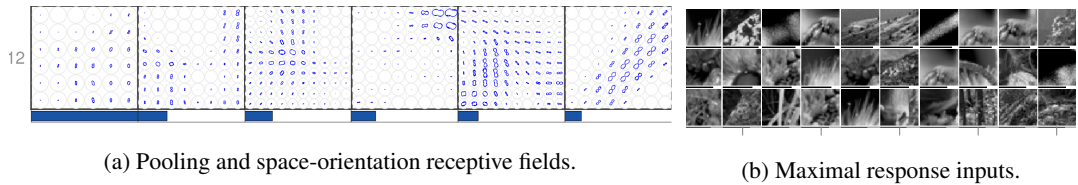

(a) Pooling and space-orientation receptive fields.

(b) Maximal response inputs.

Figure 29: Unit 12 in the 4$^{\text{th}}$ layer, visualised as in Figure 18.

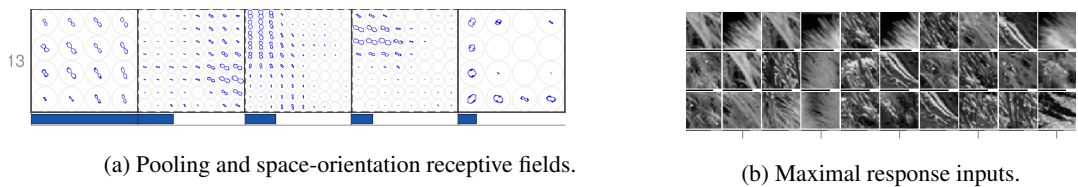

(a) Pooling and space-orientation receptive fields.

(b) Maximal response inputs.

Figure 30: Unit 13 in the 4$^{\text{th}}$ layer, visualised as in Figure 18.

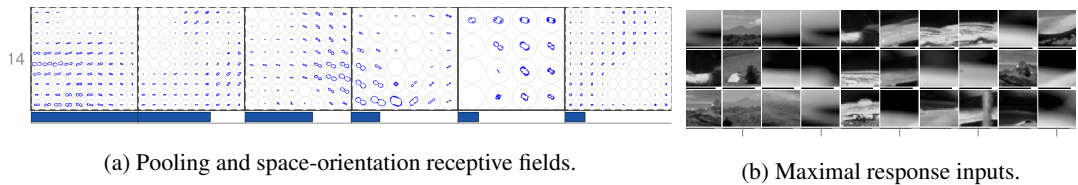

(a) Pooling and space-orientation receptive fields.

(b) Maximal response inputs.

Figure 31: Unit 14 in the 4$^{\text{th}}$ layer, visualised as in Figure 18.

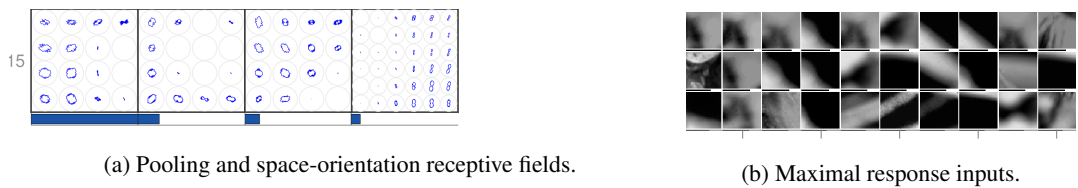

(a) Pooling and space-orientation receptive fields.

(b) Maximal response inputs.

Figure 32: Unit 15 in the 4$^{\text{th}}$ layer, visualised as in Figure 18.

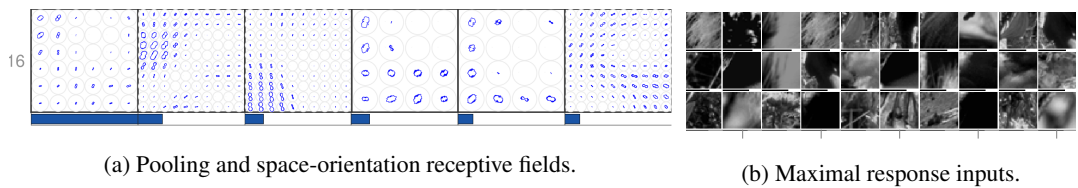

(a) Pooling and space-orientation receptive fields.

(b) Maximal response inputs.

Figure 33: Unit 16 in the 4$^{\text{th}}$ layer, visualised as in Figure 18.

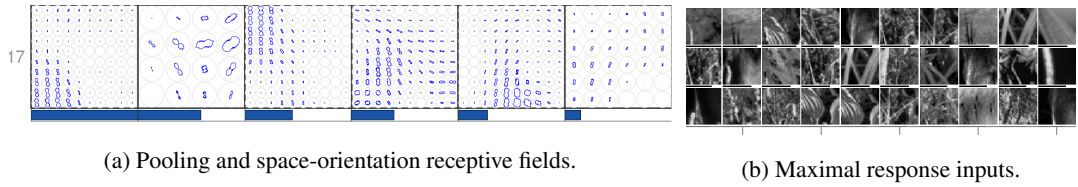

(a) Pooling and space-orientation receptive fields.

(b) Maximal response inputs.

Figure 34: Unit 17 in the 4$^{\text{th}}$ layer, visualised as in Figure 18.

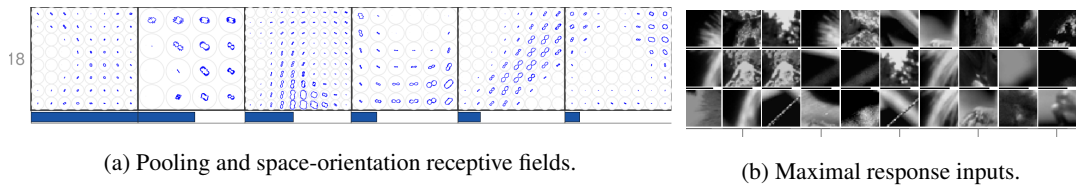

(a) Pooling and space-orientation receptive fields.

(b) Maximal response inputs.

Figure 35: Unit 18 in the 4$^{\text{th}}$ layer, visualised as in Figure 18.

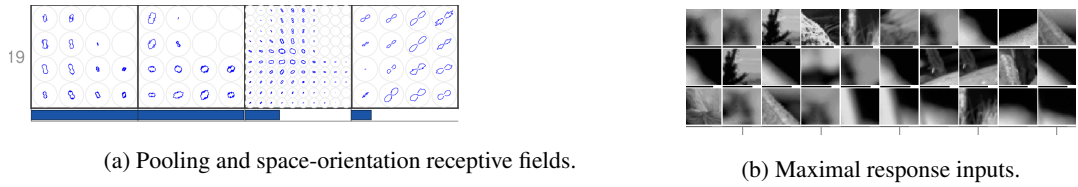

(a) Pooling and space-orientation receptive fields.

(b) Maximal response inputs.

Figure 36: Unit 19 in the 4$^{\text{th}}$ layer, visualised as in Figure 18.

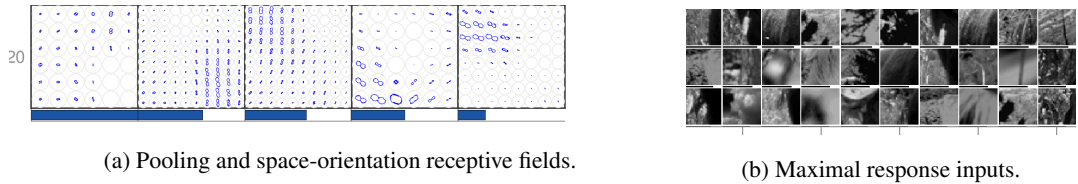

(a) Pooling and space-orientation receptive fields.

(b) Maximal response inputs.

Figure 37: Unit 20 in the 4$^{\text{th}}$ layer, visualised as in Figure 18.

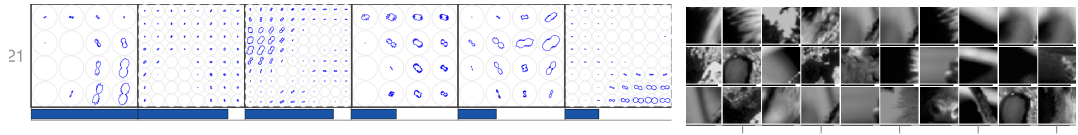

(a) Pooling and space-orientation receptive fields.

(b) Maximal response inputs.

Figure 38: Unit 21 in the 4<sup>th</sup> layer, visualised as in Figure 18.

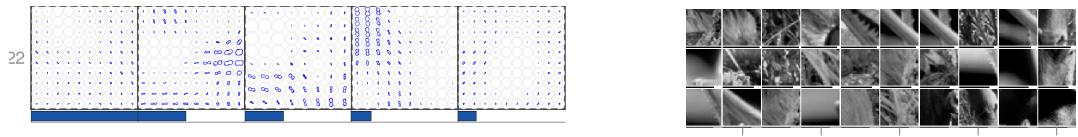

(a) Pooling and space-orientation receptive fields.

(b) Maximal response inputs.

Figure 39: Unit 22 in the 4<sup>th</sup> layer, visualised as in Figure 18.

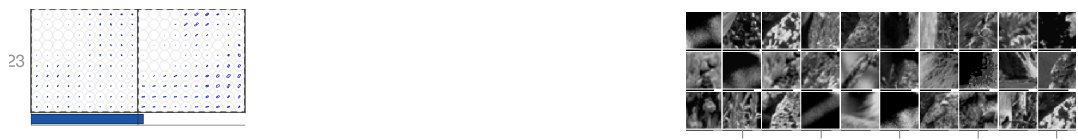

(a) Pooling and space-orientation receptive fields.

(b) Maximal response inputs.

Figure 40: Unit 23 in the 4<sup>th</sup> layer, visualised as in Figure 18.

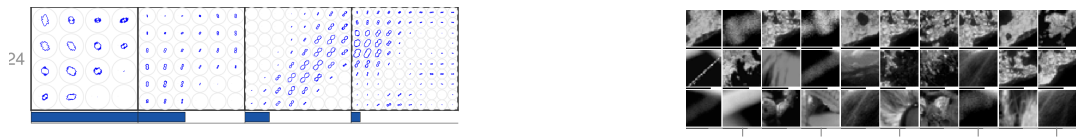

(a) Pooling and space-orientation receptive fields.

(b) Maximal response inputs.

Figure 41: Unit 24 in the 4<sup>th</sup> layer, visualised as in Figure 18.

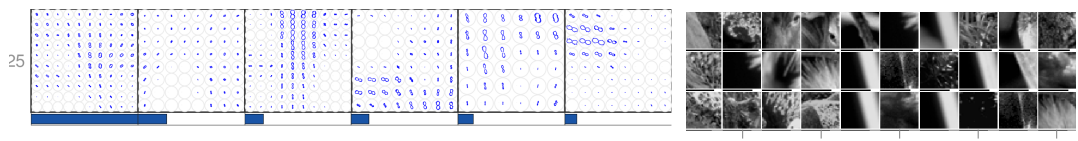

(a) Pooling and space-orientation receptive fields.

(b) Maximal response inputs.

Figure 42: Unit 25 in the 4<sup>th</sup> layer, visualised as in Figure 18.

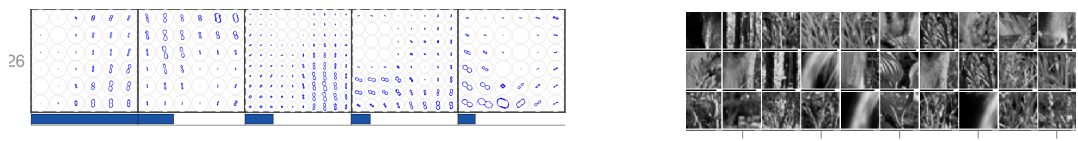

(a) Pooling and space-orientation receptive fields.

(b) Maximal response inputs.

Figure 43: Unit 26 in the 4<sup>th</sup> layer, visualised as in Figure 18.

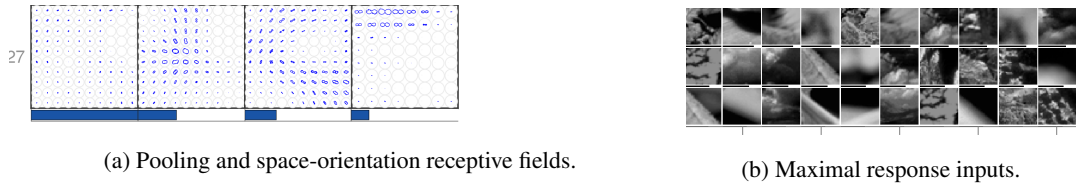

(a) Pooling and space-orientation receptive fields.

(b) Maximal response inputs.

Figure 44: Unit 27 in the 4<sup>th</sup> layer, visualised as in Figure 18.

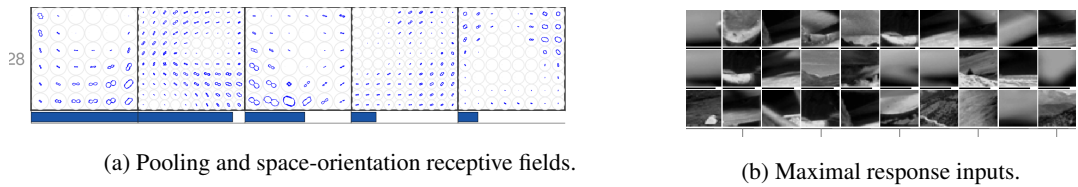

(a) Pooling and space-orientation receptive fields.

(b) Maximal response inputs.

Figure 45: Unit 28 in the 4<sup>th</sup> layer, visualised as in Figure 18.

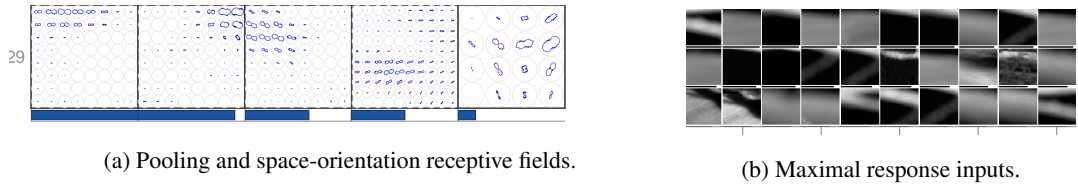

(a) Pooling and space-orientation receptive fields.

(b) Maximal response inputs.

Figure 46: Unit 29 in the 4<sup>th</sup> layer, visualised as in Figure 18.

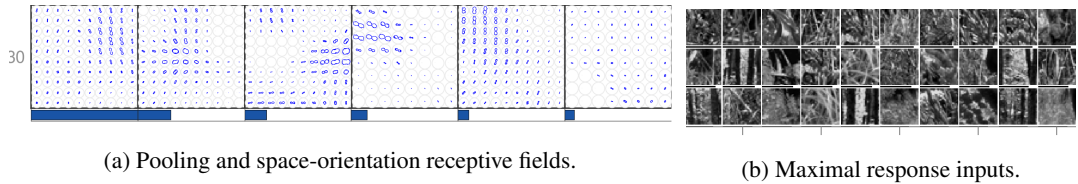

(a) Pooling and space-orientation receptive fields.

(b) Maximal response inputs.

Figure 47: Unit 30 in the 4<sup>th</sup> layer, visualised as in Figure 18.